\documentclass[10pt,twoside,english]{article}
\usepackage[T1]{fontenc}
\usepackage[latin9]{inputenc}
\usepackage{float}
\usepackage{amsmath}
\usepackage{graphicx}
\usepackage{amssymb}
\makeatletter

\providecommand{\tabularnewline}{\\}
\floatstyle{ruled}
\newfloat{algorithm}{tbp}{loa}
\floatname{algorithm}{Algorithm}

\usepackage{jmlr2e}
\usepackage{graphicx,psfrag,color,amsmath}
\usepackage{subfigure}
\usepackage{algorithmic}
\usepackage{algorithm}

\let\oldmarginpar\marginpar
\renewcommand\marginpar[1]{\-\oldmarginpar[\raggedleft\footnotesize #1]%
{\raggedright\footnotesize #1}}

\newcommand{\reals}{\mathbb{R}}
\newcommand{\dsp}{\displaystyle}
\newcommand{\ones}{\mathbf 1}
\newcommand{\zeros}{\mathbf 0}
\jmlrheading{0}{0}{0}{0}{0}{Laurent El Ghaoui, Vivian Viallon and Tarek Rabbani}
\ShortHeadings{Safe Feature Elimination}{El Ghaoui et al}
\firstpageno{1}
\usepackage{graphicx}

\makeatother

\usepackage{babel}

\begin{document}

\title{Safe Feature Elimination for the LASSO \\ and Sparse Supervised
Learning Problems}


\author{\name Laurent El Ghaoui \email elghaoui@eecs.berkeley.edu \\
       \name Vivian Viallon \email viallon@eecs.berkeley.edu \\
       \name Tarek Rabbani \email trabbani@berkeley.edu \\
       \addr Department of EECS\\
       University of California\\
       Berkeley, CA 94720-1776, USA}

\editor{ }

\maketitle

\begin{abstract}
We describe a fast method to eliminate features (variables) in  $l_1$-penalized
least-square regression (or LASSO) problems. The elimination of features leads
to a potentially substantial reduction in running time, especially for large
values of the penalty parameter. Our method is not heuristic: it only eliminates
features that are guaranteed to be absent after solving the LASSO problem. The
feature elimination step is easy to parallelize and can test each feature for
elimination independently. Moreover, the computational effort of our method is
negligible compared to that of solving the LASSO problem - roughly it is the
same as single gradient step. Our method extends the scope of existing LASSO 
algorithms to treat larger data sets, previously out of their reach. We show
how our method can be extended to general $l_1$-penalized convex problems and
 present preliminary results for the Sparse Support Vector Machine and Logistic
Regression problems. 
\end{abstract}


\section{Introduction}

``Sparse'' classification or regression problems, which involve
an $\ell_{1}-\text{norm}$ regularization has attracted a lot of interest
in the statistics~\citep{tibshirani1996regression}, signal processing~\citep{chen2001atomic},
and machine learning communities. The $\ell_{1}$ regularization leads
to sparse solutions, which is a desirable property to achieve model
selection, or data compression. For instance, consider the problem
of $\ell_{1}$-regularized least square regression commonly referred
to as the LASSO~\citep{tibshirani1996regression}. In this context,
we are given a set of $m$ observations $a_{i}\in\reals^{n},\, i=1,\ldots ,m$
and a response vector $y\in\reals^{m}$ . Denoting by $X=\left(a_{1},\ldots ,a_{m}\right)^{T}\in\reals^{m\times n}$
the feature matrix of observations, the LASSO problem is given by
\begin{equation}
{\cal P}(\lambda)\;:\;\phi(\lambda):=\min_{w}\frac{1}{2}\left\Vert Xw-y\right\Vert _{2}^{2}+\lambda\left\Vert w\right\Vert _{1},\label{eq:LASSO}\end{equation}
where $\lambda$ is a regularization parameter and $w\in\reals^{n}$
is the optimization variable. For large enough values of $\lambda$,
any solution $w^{\star}\in\reals^{n}$ of (\ref{eq:LASSO}) is typically
sparse, i.e. $w^{\star}$ has few entries that are non-zero, and therefore
identifies the features in $X$ (columns of $X$) that are useful
to predict $y$.

Several efficient algorithms have been
developed for the LASSO problem, including \cite{lars,boyd_linear,ParkHastie,Homotopy,Friedman_Pathwise,becker2010templates,FHT:10}
and references therein. However, the complexity of these algorithms,
when it is known, grows fast with the number of variables. While the
LASSO problem is particularly appealing in presence of very high-dimensional
problems, the available algorithms can be quite slow in such contexts.
In some applications, the feature matrix is so big that it can not
even be loaded and LASSO solvers cannot be used at all. Hence it is
of paramount interest to be able to efficiently eliminate features
in a pre-processing step, in order to reduce dimensionality and solve
the optimization problem on a reduced matrix.

Assume that a sparse solution exists to (\ref{eq:LASSO}) and that
we were able to identify $e$ zeros of $w^{\star}$ \textbf{a priori}
to solving the LASSO problem. Identifying $e$ zeros in
$w^{\star}$ a priori to solving (\ref{eq:LASSO}) is equivalent to
removing $e$ features (columns) from the feature matrix $X$. If $e$ is large,
we can obtain $w^{\star}$ by solving (\ref{eq:LASSO}) with a ``small''
feature matrix $X$. 

In this paper we propose a ``safe'' feature elimination (SAFE) method that can
identify zeros in the solution $w^{\star}$ a priori to solving the
LASSO problem. Once the zeros are identified we can safely remove
the corresponding features and then solve the LASSO problem (\ref{eq:LASSO})
on the reduced feature matrix. 

Feature selection methods are often used to accomplish dimensionality
reduction, and are of utmost relevance for data sets of massive dimension,
see for example \cite{Fan_Lv_Review}. These methods, when used as
a pre-processing step, have been referred to in the literature as
{\em screening} procedures \citep{Fan_Lv_Review,Fan_Lv}. They
typically rely on univariate models to score features, independently
of each other, and are usually computationally fast. Classical procedures
are based on correlation coefficients, two-sample $t$-statistics
or chi-square statistics \citep{Fan_Lv_Review}; see also \cite{BNS}
and the references therein for an overview in the specific case of
text classification. Most screening methods might remove features
that could otherwise have been selected by the regression or classification
algorithm. However, some of them were recently shown to enjoy the
so-called ``sure screening'' property \citep{Fan_Lv}: under some
technical conditions, no relevant feature is removed, with probability
tending to one.

Screening procedures typically ignore the specific classification
task to be solved after feature elimination. In this paper, we propose
to remove features based on the supervised learning problem considered,
that is on both the structure of the loss function and the problem
data. While we focus mainly on the LASSO problem here, we provide results 
for a large class of convex classification or regression
problems. The features are eliminated according to a sufficient, in
general conservative, condition, which we call SAFE (for SAfe Feature
Elimination). With SAFE, we never remove features unless they are
{\em guaranteed} to be absent if one were to solve the full-fledged
classification or regression problem.

An interesting fact is that SAFE becomes extremely aggressive at removing
features for large values of the penalty parameter $\lambda$. The
specific application we have in mind involves large data sets of text
documents, and sparse matrices based on occurrence, or other score,
of words or terms in these documents. We seek extremely sparse optimal
coefficient vectors, even if that means operating at values of the
penalty parameter that are substantially larger than those dictated
by a pure concern for predictive accuracy. The fact that we need to
operate at high values of this parameter opens the hope that, at least
for the application considered, the number of features eliminated
by using our fast test is high enough to allow a dramatic reduction
in computing time and memory requirements. Our experimental results
indicate that for many of these data sets, we do observe a dramatic
reduction in the number of variables, typically by an order of magnitude
or more. The method has two main advantages: for medium- to large-sized problem, it enables to reduce the computational time. More importantly, SAFE allows to tackle problems that are too huge to be even loaded in memory, thereby expanding the reach of current algorithms
                                      
The paper is organized as follows. In section~\ref{sec:Proposed-SAFE-method},
we derive the SAFE method for the LASSO problem. In
section~\ref{sec:using_safe}, we illustrate the use of SAFE and detail some
relevant algorithms.  In section \ref{sec:SAFE-for-general}, we extend the
results of SAFE to general convex problems and derive preliminary SAFE results
for the Sparse Support Vector Machine and Logistic regression problems. In
section ~\ref{s:num}, we experiment the SAFE for LASSO method on synthetic data
and on data derived from text classification sources. Numerical results
demonstrate that SAFE provides a substantial reduction in problem size, and, as
a result, it enables the LASSO algorithms to run faster and solve huge problems
originally out of their reach.

\paragraph{Notation.}

We use $\ones$ and $\zeros$ to denote a vector of ones and zeros,
with size inferred from context, respectively. For a scalar $a$,
$a_{+}$ denotes the positive part of $a$. For a vector $a$, this
operation is component-wise, so that $\ones^{T}a_{+}$ is the sum
of the positive elements in $a$. We take the convention that a sum
over an empty index sets, such as $\sum_{i=1}^{k}a_{i}$ with $k\le0$,
is zero.

\section{The SAFE method for the LASSO\label{sec:Proposed-SAFE-method}}
The SAFE method crucially relies on duality and optimality conditions. We begin by reviewing the appropriate facts.

\subsection{Dual problem and optimality conditions for the LASSO\label{sec:Dual-Problem-and}}

\begin{figure}
\subfigure[]{\label{subfig:geo_interp_a}\includegraphics[width=8cm]{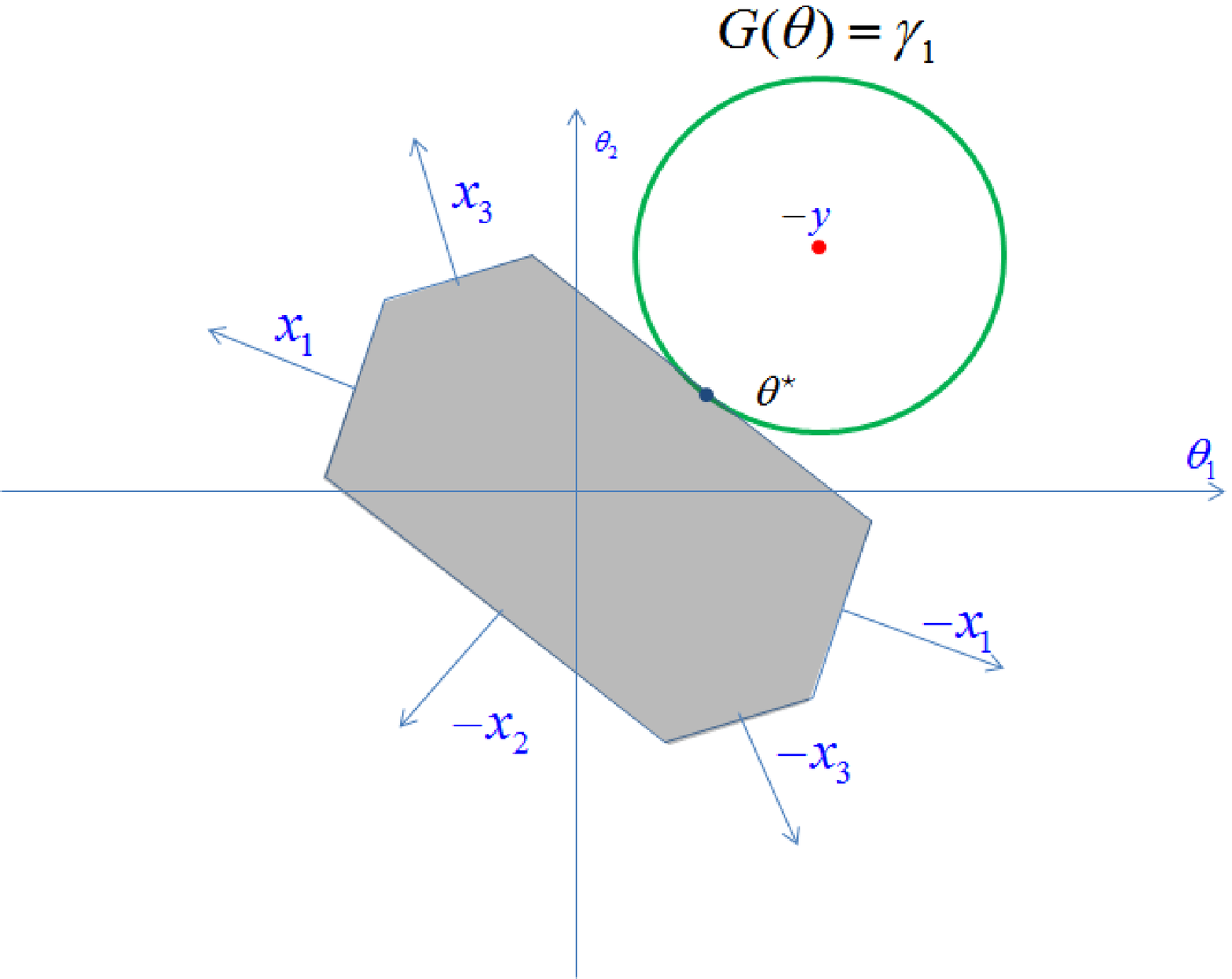}}\hfill{}\subfigure[]{\label{subfig:geo_interp_b}\includegraphics[width=8cm]{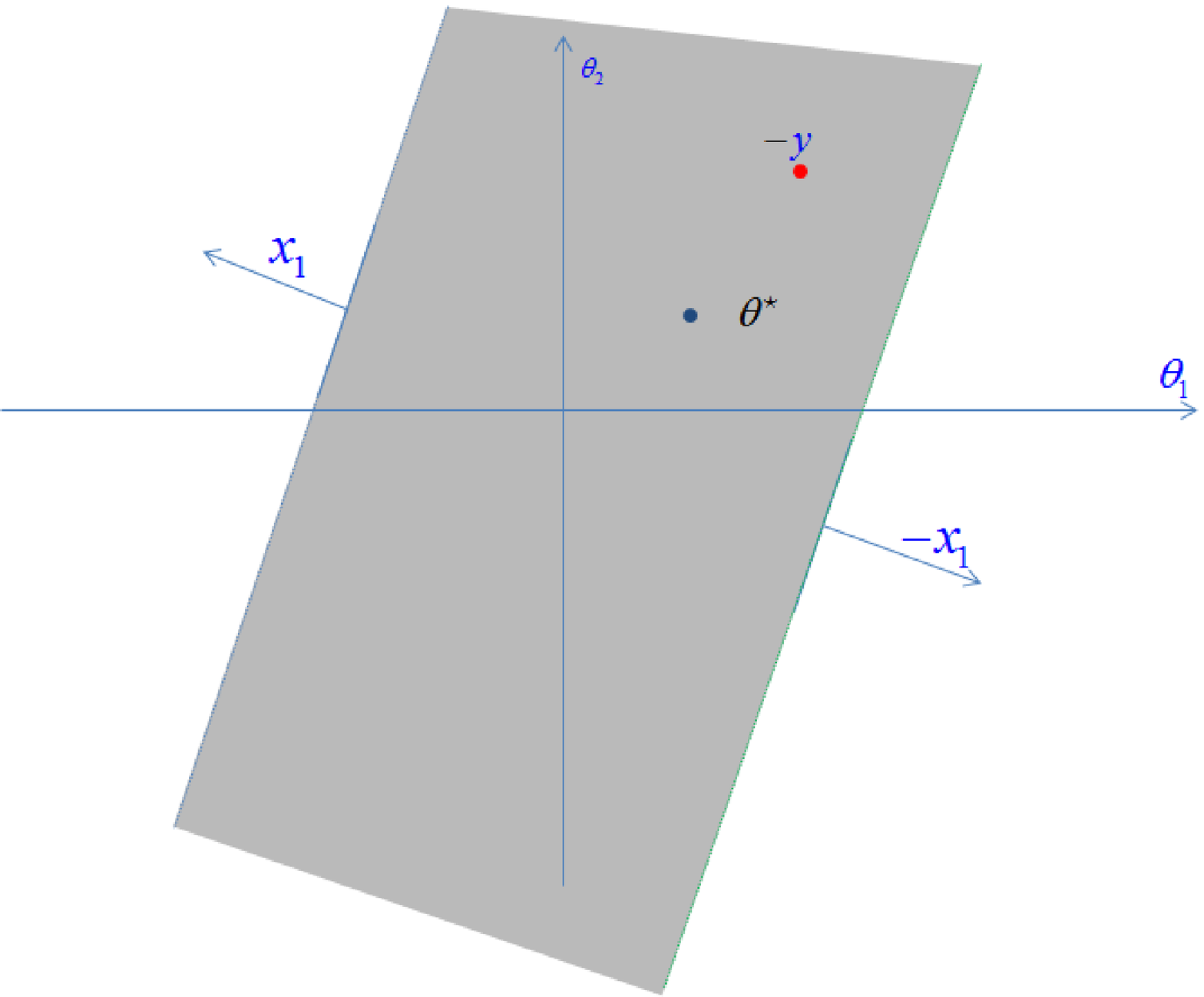}}

\caption{\label{fig:geo_interp} Geometry of the dual problem ${\cal D}(\lambda)$.
(a) Feasibility set of the dual problem. The grey shaded polytope
shows the feasibility set of ${\cal D}(\lambda)$. The feasibility
set is the intersection of $n$ slabs in the dual space corresponding
to the $n$ features $x_{k}$, $k=1,\ldots ,n$. The level set $G(\theta)=\gamma_{1}$,
where $\gamma_{1}=G(\theta^{\star})$, corresponds to the optimal
value of the dual function and is tangent to the feasibility set at
the dual optimal point $\theta^{\star}$. (b) Geometry of the inequality
test in (\ref{eq:opt_sec2}). The grey shaded region is the slab corresponding
to feature $x_{k}$, i.e. $\left\{
\theta\:|\:\left|\theta^{T}x_{k}\right|\leq\lambda\right\} $.
The test $\left|\theta^{\star T}x_{k}\right|<\lambda$ is a strict
inequality when the point $\theta^{\star}$ is in the interior of
the slab defined by the feature $x_{k}$. Thus if the dual optimal
point is inside a slab defined by feature $x_{k}$, by optimality
condition (\ref{eq:opt_sec2}) the $k$-th entry of the primal optimal
solution $w^{\star}$ is zero, i.e. $(w^{\star})_{k}=0$. 
}
\end{figure}

A dual to the LASSO problem (\ref{eq:LASSO})~\citep{boyd_linear} can be written as
\begin{equation}
{\cal D}(\lambda)\;:\;\phi(\lambda):=\max_{\theta}\: G(\theta)\::\:\left|\theta^{T}x_{k}\right|\leq\lambda,\, k=1,\ldots ,n,\label{eq:dualLASSO}\end{equation}
with $x_{k}\in\reals^{m},\, k=1,\ldots ,n$, the $k$-th column of $X$
and $G(\theta)=\frac{1}{2}\left\Vert y\right\Vert_{2}^{2}-\frac{1}{2}\left\Vert \theta+y\right\Vert _{2}^{2}$.
In this context, we call ${\cal P}(\lambda)$ the primal problem,
$w$ the primal variable, and $w^{\star}$ a primal optimal point.
The dual problem ${\cal D}(\lambda)$ is a convex optimization problem
with dual variable $\theta\in\reals^{m}$. We call $\theta$ dual
feasible when it satisfies the constraints in ${\cal D}(\lambda)$.
Figure \ref{subfig:geo_interp_a} shows the geometry of the feasibility
set in the dual space. The quantity $G(\theta)$ gives a lower bound
on the optimal value $\phi(\lambda)$ for any dual feasible point
$\theta$, i.e. $G(\theta)\leq\phi(\lambda),\;\left|\theta^{T}x_{k}\right|\leq\lambda,\, k=1,\ldots ,n$.
For the LASSO problem (\ref{eq:LASSO}) strong duality holds and the
optimal value of ${\cal D}(\lambda)$ achieves $\phi(\lambda)$ at
$\theta^{\star}$ the solution of (\ref{eq:dualLASSO}) or the dual
optimal point. Furthermore, the following relation holds at optimum:
$\theta^{\star}=Xw^{\star}-y$.

We consider the dual problem ${\cal D}(\lambda)$ because of an important
property that helps us derive our SAFE method. Assuming $w^{\star}$
is sparse, knowledge of $\theta^{\star}$ allows us to identify the
zeros in $w^{\star}$ by checking the optimality condition \citep{BV:04}:
\begin{equation}
\left|\theta^{\star T}x_{k}\right|<\lambda\Rightarrow\left(w^{\star}\right)_{k}=0.\label{eq:opt_sec2}\end{equation}

Figure \ref{subfig:geo_interp_b} illustrates the geometric interpretation
of the inequality test $\left|\theta^{\star T}x_{k}\right|<\lambda$
in (\ref{eq:opt_sec2}).

\subsection{Basic idea}
The basic idea behind SAFE is to use the optimality condition (\ref{eq:opt_sec2})
with $\theta^{\star}$ in the inequality test replaced by a set $\Theta$
that contains the dual optimal point, i.e. $\left|\theta^{T}x_{k}\right|<\lambda,\;\forall\theta\in\Theta$
and $\theta^{\star}\in\Theta$. If the inequality test holds for the
whole set $\Theta$, then the $k$-th entry of $w^{\star}$ is zero,
$(w^{\star})_{k}=0$. 

In the following sections, we show how to construct the set $\Theta$ using
optimality conditions of the dual problem, and derive the corresponding SAFE test. 

In our derivation, we assume that we have knowledge of a solution
$w_{0}^{\star}$ of ${\cal P}(\lambda_{0})$ for some $\lambda_{0}$,
and we seek to apply SAFE for ${\cal P}(\lambda)$ with $\lambda\leq\lambda_{0}$.
By default, we can choose $\lambda_{0}$ to be large enough for $w_{0}^{\star}$
to be identically zero. To find such a $\lambda_{0}$, we substitute
$w_{0}^{\star}=0$ in (\ref{eq:LASSO}) to obtain $\phi(\lambda_{0})=\frac{1}{2}\left\Vert y\right\Vert _{2}^{2}$.
By strong duality, ${\cal D}(\lambda_{0})$ achieves a value of $\phi(\lambda_{0})=\frac{1}{2}\left\Vert y\right\Vert _{2}^{2}$
at the unique solution $\theta_{0}^{\star}=-y$. The point $\theta_{0}^{\star}$
is a dual feasible point and satisfies the constraints $\lambda_{0}\geq\left|(-y)^{T}x_{k}\right|,\: k=1,\ldots ,n.$
Note that $\lambda_{0}$ is not uniquely defined but we choose the
smallest value above which $w_{0}^{\star}=0$, that is $\lambda_{0}=\max_{1\le j\le n}\:|y^{T}x_{j}|=\|X^{T}y\|_{\infty}$.

\subsection{Constructing $\Theta$\label{sub:Constructing}}

We start by finding a set $\Theta$ that contains the dual optimal
point $\theta^{\star}$ of ${\cal D}(\lambda)$. We express $\Theta$
as the intersection of two sets $\Theta_{1}$ and $\Theta_{2}$, where
each set corresponds to different optimality conditions. 

We construct
$\Theta_{1}$ using the optimality condition of ${\cal D}(\lambda)$:
$\theta^{\star}$ is a dual optimal point if $G(\theta^{\star})\geq G(\theta)$
for all dual feasible points $\theta$. Let $\theta_{s}$ be a dual
feasible point to ${\cal D}(\lambda)$, and $\gamma:=G(\theta_{s})$.
Obviously $G(\theta^{\star})\geq\gamma$ and the set $\Theta_{1}:=\left\{ \theta\:\mid\: G(\theta)\geq\gamma\right\} $
contains $\theta^{\star}$, i.e. $\theta^{\star}\in\Theta_{1}$.

One way to obtain a lower bound $\gamma$ is by dual scaling. We set $\theta_{s}$
to be a scaled feasible dual point in terms of $\theta_{0}^{\star}$,
$\theta_{s}:=s\theta_{0}^{\star}$
with $s\in\reals$ constrained so that $\theta_{s}$ is a dual feasible
point for ${\cal D}(\lambda)$, that is, $\|X^{T}\theta_{s}\|_{\infty}\le\lambda$
or $|s|\le\lambda/\lambda_{0}$. We then set $\gamma$ according to
the convex optimization problem: \[
\gamma=\max_{s}\:\left\{ G(s\theta_{0}^{\star})~:~|s|\le\frac{\lambda}{\lambda_{0}}\right\} =\max_{s}\:\left\{ \beta_{0}s-\frac{1}{2}s^{2}\alpha_{0}~:~|s|\le\frac{\lambda}{\lambda_{0}}\right\} ,\]
 with $\alpha_{0}:=\theta_{0}^{\star T}\theta_{0}^{\star}>0$, $\beta_{0}:=|y^{T}\theta_{0}^{\star}|$.
We obtain \begin{equation}
\gamma=\frac{\beta_{0}^{2}}{2\alpha_{0}}\left(1-\left(1-\frac{\alpha_{0}}{\beta_{0}}\frac{\lambda}{\lambda_{0}}\right)_{+}^{2}\right).\label{eq:gamma}\end{equation}

We construct $\Theta_{2}$ by applying a first order optimality condition
on ${\cal D}(\lambda_{0})$: $\theta_{0}^{\star}$ is a dual optimal
point if $g^{T}(\theta_{0}-\theta_{0}^{\star})\leq0$ for every dual
 point $\theta_{0}$ that is feasible for ${\cal D}(\lambda_{0})$, where $g:=\nabla G(\theta_{0}^{\star})=\theta_{0}^{\star}+y$.
For $\lambda\leq\lambda_{0}$, any dual point $\theta$ feasible for
${\cal D}(\lambda)$ is also dual feasible for ${\cal D}(\lambda_{0})$
($|\theta^{T}x_{k}|\le\lambda\leq\lambda_{0}\;\; k=1,\ldots,n$).
Since $\theta^{\star}$ is dual feasible for ${\cal D}(\lambda_{0})$,
we conclude $\theta^{\star}\in\Theta_{2}:=\left\{ \theta\mid g^{T}(\theta-\theta_{0}^{\star})\leq0\right\}$.

\begin{figure}
\subfigure[]{\label{fig:Theta}\includegraphics[width=8cm]{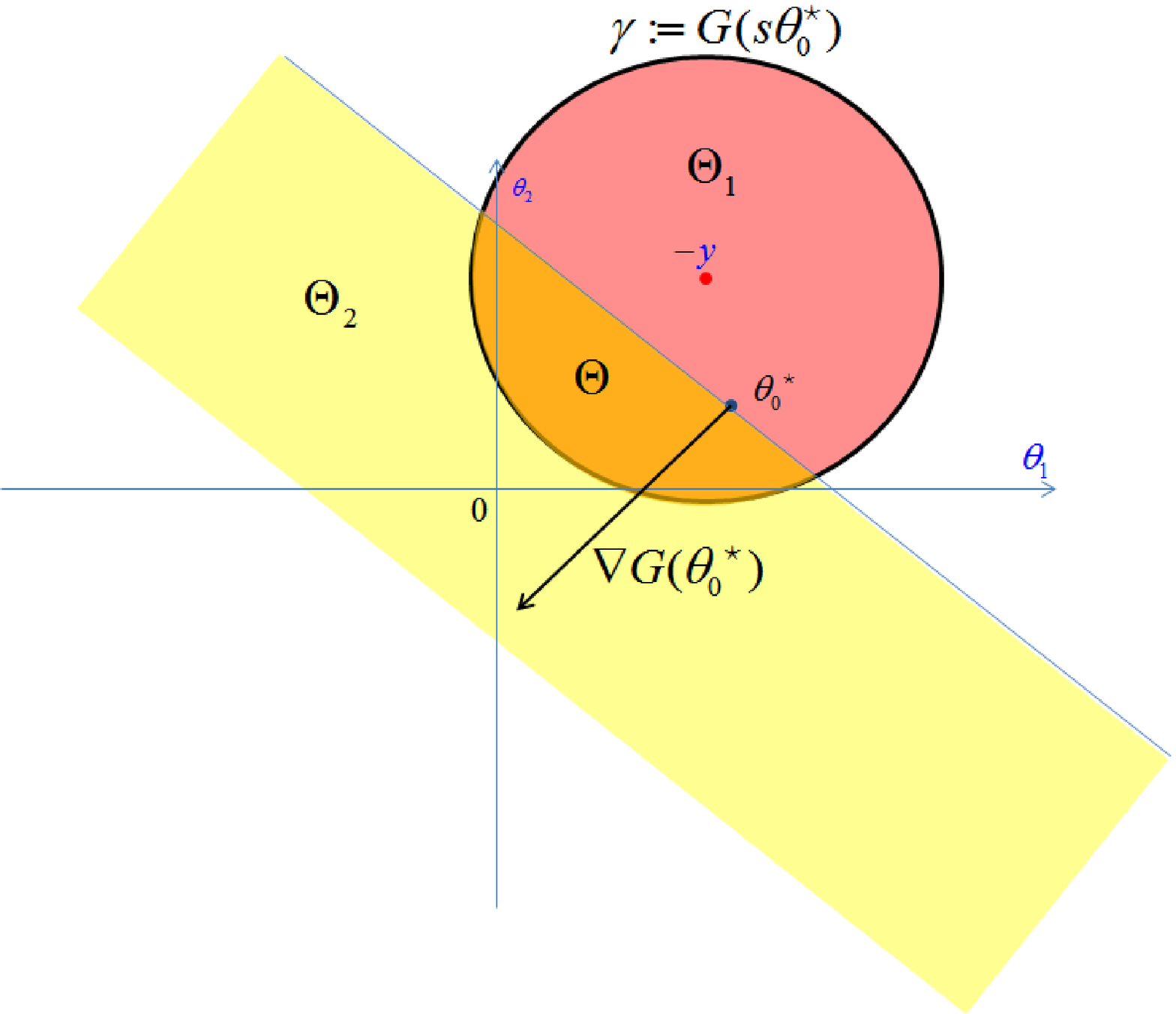}}\hfill{}\subfigure[]{\label{subfig:inequality_test}\includegraphics[width=8cm]{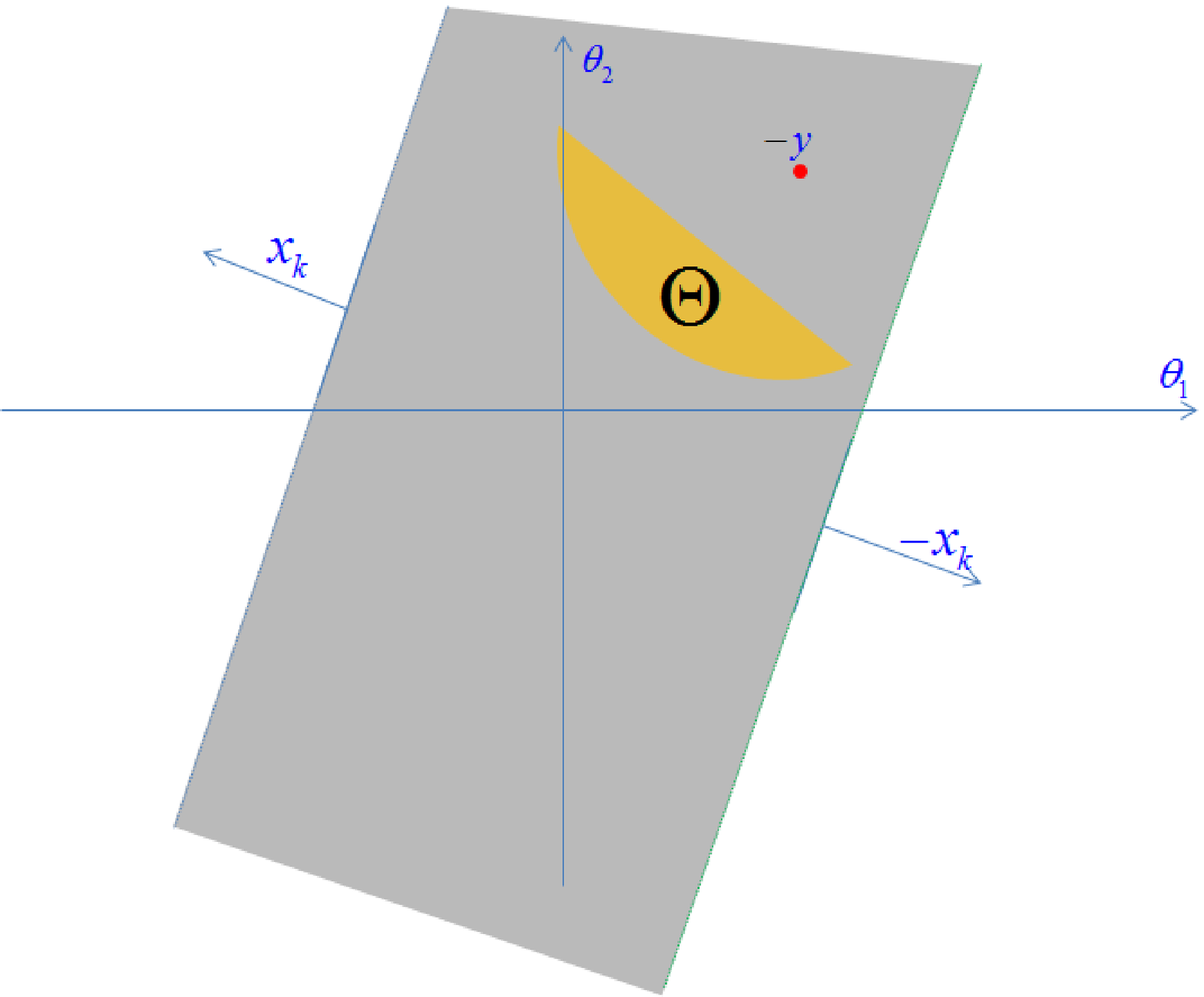}}

\caption{(a) Sets containing $\theta^{\star}$ in the dual space. The set $\Theta_{1}:=\left\{ \theta\:\mid\: G(\theta)\geq\gamma\right\} $
shown in red corresponds to a ball in the dual space with center $-y$.
The set $\Theta_{2}:=\left\{ \theta\mid g^{T}(\theta-\theta_{0}^{\star})\leq0\right\} $
with $g:=\nabla G(\theta_{0}^{\star})$ shown in yellow corresponds
to a half space with supporting hyperplane passing through $\theta_{0}^{\star}$
and normal to $\nabla G(\theta_{0}^{\star})$. The set $\Theta=\Theta_{1}\cap\Theta_{2}$
shown in orange contains the dual optimal point $\theta^{\star}$.
(b) Geometry of the inequality test $\left|\theta^{T}x_{k}\right|<\lambda,\;\forall\theta\in\Theta$.
The grey shaded region is the slab corresponding to feature $x_{k}$,
i.e. $\left\{ \theta\:|\:\theta^{T}x_{k}\leq\lambda\right\} $. The
test $\left|\theta^{T}x_{k}\right|<\lambda,\;\forall\theta\in\Theta$
is a strict inequality when the entire set $\Theta$ (shown in orange)
is inside the slab defined by the feature $x_{k}$. In such case,
the dual optimal point $\theta^{\star}\in\Theta$ is also inside the
slab and by (\ref{eq:opt_sec2}) we conclude $(w^{\star})_{k}=0$.
\label{fig:SAFE_test}}

\end{figure}

Figure \ref{fig:Theta} shows the geometry of $\Theta_{1}$, $\Theta_{2}$
and $\Theta$ in the dual space; Figure \ref{subfig:inequality_test}
shows the geometric interpretation of the inequality test when it
is applied to the set $\Theta$.

\subsection{SAFE-LASSO theorem\label{sub:SAFE-method}}

Our criterion to identify the $k$-th zero in $w^{\star}$ and thus
remove the $k$-th feature (column) from the feature matrix $X$ in
problem ${\cal P}(\lambda)$ becomes \begin{equation}
\lambda>\left|\theta^{T}x_{k}\right|=\text{max}(\theta^{T}x_{k},-\theta^{T}x_{k})\;:\;\theta\in\Theta.\label{eq:test1}\end{equation}

An equivalent formulation of condition (\ref{eq:test1}) is \[
\lambda>\text{max}(P(\gamma,x_{k}),P(\gamma,-x_{k})),\]
where $P(\gamma,x_{k})$ is the optimal value of a convex optimization
problem with constraints $\theta\in\Theta_{1}$ and $\theta\in\Theta_{2}$: 

\begin{equation}
P(\gamma,x_{k}):=\;\max_{\theta}\, x_{k}^{T}\theta\;:\; G(\theta)\geq\gamma,\; g^{T}\left(\theta-\theta_{0}^{\star}\right)\geq0.\label{eq:testproblem}
\end{equation}
It turns out that the above problem is simple enough to admit a closed-form solution (see Appendix~\ref{app:p-gamma}). The resulting test can be summarized as follows.


{\bf Theorem (SAFE-LASSO)} { \it Consider the LASSO problem
${\cal P}(\lambda)$ in~(\ref{eq:LASSO}). Let $\lambda_{0}\ge\lambda$ be a value
for which an optimal solution $w_{0}^{\star}\in\reals^{n}$ is known. Denote by
$x_{k}$ the $k$-th feature (column) of the matrix $X$. Define
\begin{equation}
{\cal E}=\left\{ k\:|\:\lambda>\max(P(\gamma,x_{k}),P(\gamma,-x_{k})\right\} ,\label{eq:test-LASSO-lambda0-recursive}\end{equation}
where
\begin{equation}
P(\gamma,x_{k})=\begin{cases}
\theta_{0}^{\star T}x_{k}+\Psi_{k}\tilde{D}(\gamma)\qquad & \left\Vert g\right\Vert _{2}^{2}\left\Vert x_{k}\right\Vert _{2}\geq D(\gamma)x_{k}^{T}g,\\
-y^{T}x_{k}+\left\Vert x_{k}\right\Vert _{2}D(\gamma)\qquad & \left\Vert g\right\Vert _{2}^{2}\left\Vert x_{k}\right\Vert _{2}\leq D(\gamma)x_{k}^{T}g,\end{cases}\label{eq:thmP}
\end{equation}
with
\[
\begin{array}{l}
\theta_{0}^{\star}=Xw_{0}^{\star}-y, \;\; g:=\theta_{0}^{\star}+y, \;\;
\alpha_{0}:=\theta_{0}^{\star T}\theta_{0}^{\star}, \;\;
\beta_{0}:=|y^{T}\theta_{0}^{\star}|, \;\;
\gamma:=\frac{\beta_{0}^{2}}{2\alpha_{0}}\left(1-\left(1-\frac{\alpha_{0}}{\beta_{0}}\frac{\lambda}{\lambda_{0}}\right)_{+}^{2}\right), \\
D(\gamma)=\left(\left\Vert y\right\Vert _{2}^{2}-2\gamma\right)^{1/2}, \;\;
\tilde{D}(\gamma)=\left(D(\gamma)^{2}-\left\Vert g\right\Vert _{2}^{2}\right)^{1/2}, \;\;
\Psi_{k}:=\left(\left\Vert x_{k}\right\Vert _{2}^{2}-\frac{\left(x_{k}^{T}g\right)^{2}}{\left\Vert g\right\Vert _{2}^{2}}\right)^{1/2}.	
\end{array}
\]
Then, for every index $e\in{\cal E}$, the $e$-th entry of $w^{\star}$ is zero, i.e. $\left(w^{\star}\right)_{e}=0$, and feature $x_{e}$ can be safely eliminated from $X$ a priori to
solving the LASSO problem (\ref{eq:LASSO}).} \hfill\BlackBox

When we don't have access to a solution $w_{0}^{\star}$ of ${\cal P}(\lambda_{0})$,
we can set $w_{0}^{\star}=0$ and $\lambda_{0}=\lambda_{\rm max}:=\|X^{T}y\|_{\infty}$.
In this case, the inequality test $\lambda>\text{max}(P(\gamma,x_{k}),P(\gamma,-x_{k})$
in the SAFE-LASSO theorem takes the form $\lambda>\rho_{k}\lambda_{\rm
max}$,
with 
\[
\rho_{k}=\frac{\left\Vert y\right\Vert _{2}\left\Vert x_{k}\right\Vert _{2}+|y^{T}x_{k}|}{\left\Vert y\right\Vert _{2}\left\Vert x_{k}\right\Vert _{2}+\lambda_{\rm max}}.
\]

In the case of scaled data sets, for which $\left\Vert y\right\Vert _{2}=1$
and $\left\Vert x_{k}\right\Vert _{2}=1$ for every $k$, $\rho_{k}$
has a convenient geometrical interpretation:\[
\rho_{k}=\frac{1+\left|\cos\alpha_{k}\right|}{1+\underset{1\le j\le n}{\text{max}}\:|\cos\alpha_{j}|},\]
where $\alpha_{k}$ is the angle between the $k$-th feature and the
response vector $y$. Our test then consists in eliminating features
based on how closely they are aligned with the response, \textit{relative}
to the most closely aligned feature. For scaled data sets, our test
is very similar to standard correlation-based feature selection~\citep{Fan_Lv};
in fact, for scaled data sets, the ranking of features it produces
is then exactly the same. The big difference here is that our test
is not heuristic, as it only eliminates features that are \textit{guaranteed}
to be absent when solving the full-fledged sparse supervised learning
problem.

\subsection{SAFE for LASSO with intercept problem}
\label{ss:intercept-lasso}
The SAFE-LASSO theorem can be applied to the LASSO with intercept
problem\[
{\cal P}_{\text{int}}(\lambda)\;:\;\phi(\lambda):=\min_{w,\nu}\frac{1}{2}\left\Vert Xw+\nu-y\right\Vert _{2}^{2}+\lambda\left\Vert w\right\Vert _{1},\]
 with $\nu\in\reals^{m}$ the intercept term, by using a simple transformation.
Taking the derivative of the objective function of ${\cal P}_{\text{int}}(\lambda)$
w.r.t $\nu$ and setting it to zero, we obtain $\nu=\bar{y}-\bar{X}^{T}w$
with $\bar{y}=(1/m)\ones^{T}y$, $\bar{X}=(1/m)X\ones$ and $\ones\in\reals^{m}$
the vector of ones . Using the expression of $\nu$, ${\cal P}_{\text{int}}(\lambda)$
can be expressed as\[
{\cal P}_{\text{int}}(\lambda)\;:\;\phi(\lambda):=\min_{w}\frac{1}{2}\left\Vert X_{\text{cent}}w-y_{\text{cent}}\right\Vert _{2}^{2}+\lambda\left\Vert w\right\Vert _{1},\]
 with $X_{\text{cent}}:=X-\bar{X}\ones^{T}$ and $y_{\text{cent}}=y-\bar{y}\ones$.
Thus the SAFE-LASSO theorem can be applied to ${\cal P}_{\text{int}}$
and eliminate features (columns) from $X_{\text{cent}}$ .

\subsection{SAFE for elastic net}

The elastic net problem \[
{\cal P}_{\text{elastic}}(\lambda)\;:\;\phi(\lambda):=\min_{w}\frac{1}{2}\left\Vert Xw-y\right\Vert _{2}^{2}+\lambda\left\Vert w\right\Vert _{1}+\frac{1}{2}\epsilon\left\Vert w\right\Vert _{2}^{2},\]
 can be expressed in the form of ${\cal P}(\lambda)$ by replacing
$X$ and $y$ of (\ref{eq:LASSO}) with $X_{\text{elastic}}=\left(X^{T},\sqrt{\epsilon}I\right)^{T}$
and $y_{\text{elastic}}=\left(y^{T},\mathbf{0}^{T}\right)^{T}$. This
transformation allows us to apply the SAFE-LASSO theorem on ${\cal
P}_{\text{elastic}}(\lambda)$
and eliminate features from $X_{\text{elastic}}$.

\section{Using SAFE} 
\label{sec:using_safe}
In this section we illustrate the use of SAFE and detail the relevant algorithms. 

\subsection{SAFE for reducing memory limit problems}
SAFE can extend the reach of LASSO solvers to larger size problems
than what they could originally handle. In this section, we are interested
in solving for $w_{d}^{\star}$ the solution of ${\cal P}(\lambda_{d})$
under a memory constraint of loading only $M$ features. We can compute
$w_{d}^{\star}$ by solving a sequence of problems, where each problem
has a number of features less than our memory limit $M$. We start
by finding an appropriate $\lambda$ where our SAFE method can eliminate
at least $n-M$ features, we then solve a reduced size problem with
$L_{F}\leq M$ features, where $L_{F}=\left|{\cal E}^{c}\right|$
is the number of features left after SAFE and ${\cal E}^{c}=\left\{ 1,\ldots ,n\right\} \backslash{\cal E}$
is the complement of the set ${\cal E}$ in the SAFE-LASSO theorem.
We proceed to the next stage as outlined in algorithm \ref{alg:SAFE-for-memory}.%
\begin{algorithm}[H]
\textbf{given }a feature matrix $X\in\reals^{m\times n}$, response
$y\in\reals^{m}$, penalty parameter $\lambda_{d}$ , memory limit
$M$ and LASSO solver: \verb"LASSO", i.e. $w^\star=\verb"LASSO"(X,y,\lambda)$.

\textbf{initialize} $\lambda_{0}=\|X^{T}y\|_{\infty}$, $w_{0}^{\star}=\zeros\in\reals^{n}$,

\textbf{repeat}
\begin{enumerate}
\item \textit{Use SAFE to search for a $\lambda$ with $LF\leq M$ .} Obtain
$\lambda$ and ${\cal E}$. \% $L_{F}$ is the number of features
left after SAFE and ${\cal E}$ is the set defined in the SAFE-LASSO theorem.
\item \textbf{if} $\lambda<\lambda_{d}$ \textbf{then} $\lambda=\lambda_{d}$,
apply SAFE to obtain ${\cal E}$ \textbf{end if.}
\item Compute the solution $w^{\star}$. $w^{\star}({\cal E}^{c})=\verb"LASSO"(X({\cal E}^{c},:),y,\lambda)$,
$w^{*}({\cal E})=0$; \% $w^{\star}({\cal E}^{c})$ and $X({\cal E}^{c},:)$
are the elements and columns of $w^{\star}$ and $X$ defined by the
set ${\cal E}^{c}$, respectively. ${\cal E}^{c}=\left\{ 1,\ldots ,n\right\} \backslash{\cal E}$
is the complement of the set ${\cal E}$ .
\item $\lambda_{0}:=\lambda$, $w_{0}^{\star}=w^{*}$.
\end{enumerate}
\textbf{until $\lambda_{0}=\lambda_{d}$}

\caption{SAFE for reducing memory limit problems\label{alg:SAFE-for-memory}}

\end{algorithm}

We use a bisection method to find an appropriate value of $\lambda$
for which SAFE leaves $L_{F}\in\left[M-\epsilon_{F},\: M\right]$
features, where $\epsilon_{F}$ is a number of feature tolerance.
The bisection method on $\lambda$ is outlined in algorithm \ref{alg:Bisection-method-for}.%
\begin{algorithm}[h]
\textbf{given }a feature matrix $X\in\reals^{m\times n}$, response
$y\in\reals^{m}$, penalty parameter $\lambda_{0}$ with LASSO solution
$w_{0}^{\star}$, tolerance $\epsilon_{F}>0$ and memory limit $M$.

\textbf{initialize} $l=0$, and $u=\lambda_{0}$. 

\textbf{repeat}
\begin{enumerate}
\item Set $\lambda:=\left(l+u\right)/2$.
\item Use the SAFE-LASSO theorem to obtain ${\cal E}$. 
\item Set $L_{F}=\left|{\cal E}^{c}\right|$. 
\item \textbf{if} $L_{F}>M$ \textbf{then }set $l:=\lambda$\textbf{ else}
set $u:=\lambda$ \textbf{end if }
\end{enumerate}
\textbf{until} $M-L_{F}\leq\epsilon_{F}$ and $L_{F}\leq M$.

\caption{Bisection method on $\lambda$. \label{alg:Bisection-method-for}}

\end{algorithm}

\subsection{SAFE for LASSO run-time reduction}
In some applications like \cite{GJMEYC:10}, it is of interest to
solve a sequence of problems ${\cal P}(\lambda_{1}),\:\ldots {\cal P}(\lambda_{s})$
for decreasing values of the penalty parameters, i.e. $\lambda_{1}\geq\ldots \geq\lambda_{s}$.
The computational complexities of LASSO solvers depend on the number
of features and using SAFE might result in run-time improvements.
For each problem in the sequence, we can use SAFE to reduce the number
of features a priori to using our LASSO solver as shown in algorithm
\ref{alg:Recursive-SAFE-for-1}.%
\begin{algorithm}[h]
\textbf{given }a feature matrix $X\in\reals^{m\times n}$, response
$y\in\reals^{m}$, a sequence of penalty parameters $\lambda_{s}\leq\ldots \leq\lambda_{1}\leq\|X^{T}y\|_{\infty},$
and LASSO solver: \verb"LASSO".

\textbf{initialize} $\lambda_{0}=\|X^{T}y\|_{\infty}$, $w_{0}^{\star}=\zeros\in\reals^{n}$.

\textbf{for} $i=1$ \textbf{until} $i=s$ \textbf{do}
\begin{enumerate}
\item Set $\lambda_{0}=\lambda_{i-1}$, and $\lambda=\lambda_{i}$.
\item Use the SAFE-LASSO theorem to obtain ${\cal E}$. 
\item Compute the solution $w^{\star}$. $w^{\star}({\cal E}^{c})=\verb"LASSO"(X({\cal E}^{c},:),y,\lambda)$,
$w^{*}({\cal E})=0$. \% $w^{\star}({\cal E}^{c})$ and $X({\cal E}^{c},:)$
are the elements and columns of $w^{\star}$ and $X$ defined by the
set ${\cal E}^{c}$, respectively. ${\cal E}^{c}=\left\{ 1,\ldots ,n\right\} \backslash{\cal E}$
is the complement of the set ${\cal E}$ .
\item Set $w_{0}^{\star}=w^{*}$.
\end{enumerate}
\textbf{end for }

\caption{Recursive SAFE for the Lasso\label{alg:Recursive-SAFE-for-1}}

\end{algorithm}


\section{SAFE applied to general $\ell_{1}$-regularized convex problems\label{sec:SAFE-for-general}}

The SAFE-LASSO result presented in section \ref{sub:SAFE-method}
for the LASSO problem (\ref{eq:LASSO}) can be adapted to a more general
class of $l_{1}-$ regularized convex problems. We consider the family
of problems\begin{equation}
{\cal P}(\lambda)\;:\;\phi(\lambda):=\min_{w,\,\nu}\sum_{i=1}^{m}f(a_{i}^{T}w+b_{i}v+c_{i})+\lambda\left\Vert w\right\Vert _{1},\label{eq:general-p}\end{equation}
where $f$ is a closed convex function, and non-negative everywhere,
$a_{i}\in\reals^{n}$, $i=1,\ldots,m$, $b,c\in\reals^{m}$ are given.
The LASSO problem (\ref{eq:LASSO}) is a special case of (\ref{eq:general-p})
with $f(\zeta)=(1/2)\zeta^{2}$, $a_{i}\in\reals^{n}$, $i=1,\ldots,m$
the observations, $c=-y$ is the (negative) response vector, and $b=0$.
Hereafter, we refer to the LASSO problem as ${\cal P}_{\text{LASSO}}(\lambda)$
and to the general class of $l_{1}$-regularized problems as ${\cal P}(\lambda)$.
In this section, we outline the steps necessary to derive a SAFE method
for the general problem ${\cal P}(\lambda)$. We show some preliminary
results for deriving SAFE methods when $f(\zeta)$ is the hing loss
function, $f_{\text{hi}}(\zeta)=\left(1-\zeta\right)_{+}$, and the
logistic loss function $f_{{\rm log}}(\xi)=\log(1+e^{-\xi})$.

\subsection{Dual Problem}

The first step is to devise the dual of problem~(\ref{eq:general-p}),
which is \begin{equation}
{\cal D}(\lambda)~:~~\phi(\lambda)=\max_{\theta}\: G(\theta)~:~\theta^{T}b=0,\;\;|\theta^{T}x_{k}|\le\lambda,\;\; k=1,\ldots,n,\label{eq:pb-generic-dual}\end{equation}
 where \begin{equation}
G(\theta):=c^{T}\theta-\sum_{i=1}^{m}f^{\ast}(\theta_{i})\label{eq:pb-generic-dual-fcn}\end{equation}
 with $f^{\ast}(\vartheta)=\max_{\xi}\:\xi\vartheta-f(\xi)$ the conjugate
of the loss function $f(\zeta)$, and $x_{k}$ the $k$-th column
or feature of the feature matrix $X=\left(a_{1},\ldots ,a_{m}\right)^{T}\in\reals^{m\times n}$.
$G(\theta)$ is the dual function, which is, by construction, concave.
We assume that strong duality holds and primal and dual optimal points
are attained. Due to the optimality conditions for the problem (see
\cite{BV:04}), constraints for which $|\theta^{T}x_{k}|<\lambda$
at optimum correspond to a zero element in the primal variable: $\left(w^{\star}\right)_{k}=0$,
i.e.\begin{equation}
\left|\theta^{\star T}x_{k}\right|<\lambda\Rightarrow\left(w^{\star}\right)_{k}=0.\label{eq:opt_sec6}\end{equation}

\subsection{Optimality set $\Theta$}

\label{ss:dual-scaling-generic}For simplicity, we consider only the
set $\Theta:=\left\{ \theta\:|\: G(\theta)\geq\gamma\right\} $ which
contains $\theta^{\star}$ the dual optimal point of ${\cal D}(\lambda)$.
One way to get a lower bound $\gamma$ is to find a dual point $\theta_{s}$
that is feasible for the dual problem ${\cal D}(\lambda)$, and then
set $\gamma=G(\theta_{s})$.

To obtain a dual feasible point, we can solve the problem for a higher
value $\lambda_{0}\ge\lambda$ of the penalty parameter. (In the specific
case examined below, we will see how to set $\lambda_{0}$ so that
the vector $w_{0}^{\star}=0$ at optimum.) This provides a dual point
$\theta_{0}^{\star}$ that is feasible for ${\cal D}(\lambda_{0})$,
which satisfies $\lambda_{0}=\|X\theta_{0}\|_{\infty}$. In turn,
$\theta_{0}^{\star}$ can be scaled so as to become feasible for ${\cal D}(\lambda)$.
Precisely, we set $\theta_{s}=s\theta_{0}$, with $\|X\theta_{s}\|_{\infty}\le\lambda$
equivalent to $|s|\le\lambda/\lambda_{0}$. In order to find the best
possible scaling factor $s$, we solve the one-dimensional, convex
problem \begin{equation}
\gamma(\lambda):=\max_{s}\: G(s\theta_{0})~:|s|\le\frac{\lambda}{\lambda_{0}}.\label{eq:gamma-pb}\end{equation}
 Under mild conditions on the loss function $f$, the above problem
can be solved by bisection in $O(m)$ time. By construction, $\gamma(\lambda)$
is a lower bound on $\phi(\lambda)$. We can generate an initial point
$\theta_{0}^{\star}$ by solving ${\cal P}(\lambda_{0})$ with $w_{0}=0$.
We get \[
\min_{v_{0}}\:\sum_{i=1}^{m}f(b_{i}v_{0}+c_{i})=\min_{v_{0}}\:\max_{\theta_{0}}\:\theta_{0}^{T}(bv_{0}+c)-\sum_{i=1}^{m}f^{\ast}\left(\left(\theta_{0}\right)_{i}\right)=\max_{\theta_{0}\::\: b^{T}\theta_{0}=0}\: G(\theta_{0}).\]
 Solving the one-dimensional problem above can be often done in closed-form,
or by bisection, in $O(m)$. Choosing $\theta_{0}^{\star}$ to be
any optimal for the corresponding dual problem (the one on the right-hand
side) generates a point that is dual feasible for it, that is, $G(\theta_{0}^{\star})$
is finite, and $b^{T}\theta_{0}=0$.

The point $\theta_{0}^{\star}$ satisfies all the constraints of problem
${\cal D}(\lambda)$, except perhaps for the constraint $\|X\theta\|_{\infty}\le\lambda$,
i.e. $\|X\theta_{0}^{\star}\|_{\infty}>\lambda$. Hence, if $\lambda\ge\lambda_{{\rm 0}}:=\|X\theta_{0}^{\star}\|_{\infty}$,
then $\theta_{0}^{\star}$ is dual optimal for ${\cal D}(\lambda)$
and by the optimality condition (\ref{eq:opt_sec6}) we have $w^{\star}=0$
. Note that, since $\theta_{0}^{\star}$ may not be uniquely defined,
$\lambda_{0}$ may not necessarily be the smallest value for which
$w^{\star}=0$ is optimal for the primal problem.

\subsection{SAFE method}

Assume that a lower bound $\gamma$ on the optimal value of the learning
problem $\phi(\lambda)$ is known: $\gamma\le\phi(\lambda)$. (Without
loss of generality, we can assume that $0\le\gamma\le\sum_{i=1}^{m}f(c_{i})$).
The test \[
\lambda>\text{max}(P(\gamma,x_{k}),P(\gamma,-x_{k})),\]
 allows to eliminate the $k$-th feature from the feature matrix $X$,
where $P(\gamma,x_{k})$ is the optimal value of a convex optimization
problem with two constraints: \begin{equation}
P(\gamma,x_{k}):=\max_{\theta}\:\theta^{T}x_{k}~:~G(\theta)\ge\gamma,\;\;\theta^{T}b=0.\label{eq:P-def}\end{equation}
 Since $P(\gamma,x_{k})$ decreases when $\gamma$ increases, the
closer $\phi(\lambda)$ is to its lower bound $\gamma$, the more
aggressive (accurate) our test is.

By construction, the dual function $G$ is decomposable as a sum of
functions of one variable only. This particular structure allows to
solve problem~(\ref{eq:P-def}) very efficiently, using for example
interior-point methods, for a large class of loss functions $f$.
Alternatively, we can express the problem in dual form as a convex
optimization problem with two scalar variables: \begin{equation}
P(\gamma,x_{k})=\min_{\mu>0,\:\nu}\:-\gamma\mu+\mu\sum_{i=1}^{m}f\left(\frac{\left(x_{k}\right)_{i}+\mu c_{i}+\nu b_{i}}{\mu}\right).\label{eq:P-def-dual}\end{equation}
 Note that the expression above involves the perspective of the function
$f$, which is convex (see \cite{BV:04}). For many loss functions
$f$, the above problem can be efficiently solved using a variety
of methods for convex optimization, in (close to) $O(m)$ time. We
can also set the variable $\nu=0$, leading to a simple bisection
problem over $\mu$. This amounts to ignore the constraint $\theta^{T}b=0$
in the definition of $P(\gamma,x)$, resulting in a more conservative
test. More generally, any pair $(\mu,\nu)$ with $\mu>0$ generates
an upper bound on $P(\gamma,x)$, which in turn corresponds to a valid,
perhaps conservative, test.

\subsection{SAFE for Sparse Support Vector Machine}

\label{s:hi} We turn to the sparse support vector machine classification
problem: \begin{equation}
{\cal P}_{{\rm hi}}(\lambda)~:~~\phi(\lambda):=\min_{w,v}\:\sum_{i=1}^{m}(1-y_{i}(z_{i}^{T}w+v))_{+}+\lambda\|w\|_{1},\label{eq:pb-svm-primal}\end{equation}
 where $z_{i}\in\reals^{n}$, $i=1,\ldots,m$ are the data points,
and $y\in\{-1,1\}^{m}$ is the label vector. The above is a special
case of the generic problem~(\ref{eq:general-p}), where $f(\zeta):=(1-\xi)_{+}$
is the hinge loss, $b=y$, $c=0$, and the feature matrix $X$ is
given by $X=[y_{1}z_{1},\ldots,y_{m}z_{m}]^{T}$, so that $x_{k}=[y_{1}z_{1}(k),\ldots,y_{m}z_{m}(k)]^{T}$.

We denote by ${\cal I}_{+},{\cal I}_{-}$ the set of indicies corresponding
to the positive and negative classes, respectively, and denote by
$m_{\pm}=|{\cal I}_{\pm}|$ the associated cardinalities. We define
$\underline{m}:=\min(m_{+},m_{-})$. Finally, for a generic data vector
$x$, we set $x^{\pm}=(x_{i})_{i\in{\cal I}_{\pm}}\in\reals^{m_{\pm}}$,
$k=1,\ldots,n$, the vectors corresponding to each one of the classes.

The dual problem takes the form \begin{equation}
{\cal D}_{hi}(\lambda)~:~~\phi(\lambda):=\max_{\theta}\: G_{\text{hi}}(\theta)~:~-\ones\le\theta\le0,\;\;\theta^{T}y=0,\;\;|\theta^{T}x_{k}|\le\lambda,\;\; k=1,\ldots,n.\label{eq:pb-svm-dual}\end{equation}
with $G_{\text{hi}}(\theta)=\ones^{T}\theta$.

\subsubsection{Test, $\gamma$ given}

\label{ss:test-gamma-svm} Let $\gamma$ be a lower bound on $\phi(\lambda)$.
The optimal value obtained upon setting $w=0$ in~(\ref{eq:pb-svm-primal})
is given by \begin{equation}
\min_{v}\:\sum_{i=1}^{m}(1-y_{i}v)_{+}=2\min(m_{+},m_{-}):=\gamma_{{\rm max}}.\label{eq:pb-svm-primal-w0}\end{equation}
 Hence, without loss of generality, we may assume $0\le\gamma\le\gamma_{{\rm max}}$.

The feature elimination test hinges on the quantity \begin{equation}
\begin{array}{rcl}
P_{{\rm hi}}(\gamma,x) & = & \dsp\max_{\theta}\:\theta^{T}x~:~\ones^{T}\theta\ge\gamma,\;\;\theta^{T}y=0,\;\;-\ones\le\theta\le0\\
 & = & \dsp\min_{\mu>0,\:\nu}\:-\gamma\mu+\mu\sum_{i=1}^{m}f_{{\rm hi}}\left(\dsp\frac{x_{i}-\nu y_{i}}{\mu}\right)\\
 & = & \dsp\min_{\mu>0,\:\nu}\:-\gamma\mu+\sum_{i=1}^{m}(\mu+\nu y_{i}-x_{i})_{+}.\end{array}\label{eq:P-hi-pb}\end{equation}
 In appendix~\ref{app:P-gamma-svm}, we show that for any $x$, the
quantity $P(\gamma,x)$ is finite if and only if $0\le\gamma\le\gamma_{{\rm max}}$,
and can be computed in $O(m\log m)$, or less with sparse data, via
a closed-form expression. That expression is simpler to state for
$P_{{\rm hi}}(\gamma,-x)$: \[
\begin{array}{rcl}
P_{{\rm hi}}(\gamma,-x) & = & \dsp\sum_{j=1}^{\lfloor\gamma/2\rfloor}\bar{x}_{j}-(\frac{\gamma}{2}-\lfloor\frac{\gamma}{2}\rfloor)(\bar{x}_{\lfloor\gamma/2\rfloor+1})_{+}+\sum_{j=\lfloor\gamma/2\rfloor+1}^{\underline{m}}(\bar{x}_{j})_{+},\;\;0\le\gamma\le\gamma_{{\rm max}}=2\underline{m},\\
 &  & \bar{x}_{j}:=x_{[j]}^{+}+x_{[j]}^{-},\;\; j=1,\ldots,\underline{m},\end{array}\]
 with $x_{[j]}$ the $j$-th largest element in a vector $x$, and
with the convention that a sum over an empty index set is zero. Note
that in particular, since $\gamma_{{\rm max}}=2\underline{m}$: \[
P_{{\rm hi}}(\gamma_{{\rm max}},-x)=\sum_{i=1}^{\underline{m}}(x_{[j]}^{+}+x_{[j]}^{-}).\]

\subsubsection{SAFE-SVM theorem}

\label{ss:safe-thm-svm} Following the construction proposed in section~\ref{ss:dual-scaling-generic}
for the generic case, we select $\gamma=G_{{\rm hi}}(\theta)$, where
the point $\theta$ is feasible for~(\ref{eq:pb-svm-dual}), and
can found by the scaling method outlined in section~\ref{ss:dual-scaling-generic},
as follows. The method starts with the assumption that there is a
value $\lambda_{0}\ge\lambda$ for which we know the optimal value
$\gamma_{0}$ of ${\cal P}_{{\rm hi}}(\lambda_{0})$.

\paragraph{Specific choices for $\lambda_{0},\gamma_{0}$.}

Let us first detail how we can find such values $\lambda_{0}$, $\gamma_{0}$.

We can set a value $\lambda_{0}$ such that $\lambda>\lambda_{0}$
ensures that $w=0$ is optimal for the primal problem~(\ref{eq:pb-svm-primal}).
The value that results in the least conservative test is $\lambda_{0}=\lambda_{{\rm max}}$,
where $\lambda_{{\rm max}}$ is the smallest value of $\lambda$ above
which $w=0$ is optimal: \begin{equation}
\lambda_{{\rm max}}:=\min_{\theta}\:\|X\theta\|_{\infty}~:~-\theta^{T}\ones\ge\gamma_{{\rm max}},\;\;\theta^{T}y=0,\;\;-\ones\le\theta\le0.\label{eq:lambdamax-def-svm}\end{equation}
 Since $\lambda_{{\rm max}}$ may be relatively expensive to compute,
we can settle for an upper bound $\overline{\lambda}_{{\rm max}}$
on $\lambda_{{\rm max}}$. One choice for $\overline{\lambda}_{{\rm max}}$
is based on the test derived in the previous section: we ask that
it passes for all the features when $\lambda=\overline{\lambda}_{{\rm max}}$
and $\gamma=\gamma_{{\rm max}}$. That is, we set \begin{equation}
\begin{array}{rcl}
\overline{\lambda}_{{\rm max}} & = & \dsp\max_{1\le k\le n}\:\max\left(P_{{\rm hi}}(\gamma_{{\rm max}},x_{k}),P_{{\rm hi}}(\gamma_{{\rm max}},-x_{k})\right)\\
 & = & \dsp\max_{1\le k\le n}\:\max\left(\dsp\sum_{i=1}^{\underline{m}}(x_{k}^{+})_{[j]}+(x_{k}^{-})_{[j]},\sum_{i=1}^{\underline{m}}(-x_{k}^{+})_{[j]}+(-x_{k}^{-})_{[j]}\right).\end{array}\label{eq:lambda0-def-svm}\end{equation}
 By construction, we have $\overline{\lambda}_{{\rm max}}\ge\lambda_{{\rm max}}$,
in fact: \[
\begin{array}{rcl}
\overline{\lambda}_{{\rm max}} & = & \dsp\max_{1\le k\le n}\:\max_{\theta}\:|x_{k}^{T}\theta|~:~-\theta^{T}\ones\ge\gamma_{{\rm max}},\;\;\theta^{T}y=0,\;\;-\ones\le\theta\le0\\
 & = & \dsp\max_{\theta}\:\|X\theta\|_{\infty}~:~-\theta^{T}\ones\ge\gamma_{{\rm max}},\;\;\theta^{T}y=0,\;\;-\ones\le\theta\le0,\end{array}\]
 The two values $\lambda_{{\rm max}},\overline{\lambda}_{{\rm max}}$
coincide if the feasible set is a singleton, that is, when $m_{+}=m_{-}$.
On the whole interval $\lambda_{0}\in[\lambda_{{\rm max}},\overline{\lambda}_{{\rm max}}]$,
the optimal value of problem ${\cal P}_{{\rm hi}}(\lambda_{0})$ is
$\gamma_{{\rm max}}$.

\paragraph{Dual scaling.}

The remainder of our analysis applies to any value $\lambda_{0}$
for which we know the optimal value $\gamma_{0}\in[0,\gamma_{{\rm max}}]$
of the problem ${\cal P}_{{\rm hi}}(\lambda_{0})$.

Let $\theta_{0}$ be a corresponding optimal dual point (as seen shortly,
the value of $\theta_{0}$ is irrelevant, as we will only need to
know $\gamma_{0}=\ones^{T}\theta_{0}$). We now scale the point $\theta_{0}$
to make it feasible for ${\cal P}_{{\rm hi}}(\lambda)$, where $\lambda$
($0\le\lambda\le\lambda_{0}$) is given. The scaled dual point is
obtained as $\theta=s\theta_{0}$, with $s$ solution to~(\ref{eq:gamma-pb}).
We obtain the optimal scaling $s=\lambda/\lambda_{0}$, and since
$\gamma_{0}=-\ones^{T}\theta_{0}$, the corresponding bound is \[
\gamma(\lambda)=\ones^{T}(s\theta_{0})=s\gamma_{0}=\gamma_{0}\frac{\lambda}{\lambda_{0}}.\]
 Our test takes the form \[
\lambda>\max\left(P_{{\rm hi}}(\gamma(\lambda),x),P_{{\rm hi}}(\gamma(\lambda),-x)\right).\]

Let us look at the condition $\lambda>P_{{\rm hi}}(\gamma(\lambda),-x)$:
\[
\exists\:\mu\ge0,\:\nu~:~\lambda>-\gamma(\lambda)\mu+\sum_{i=1}^{m}(\mu+\nu y_{i}+x_{i})_{+},\]
 which is equivalent to: \[
\lambda>\min_{\mu\ge0,\nu}\:\frac{\dsp\sum_{i=1}^{m}(\mu+\nu y_{i}+x_{i})_{+}}{1+(\gamma_{0}/\lambda_{0})\mu}.\]
 The problem of minimizing the above objective function over variable
$\nu$ has a closed-form solution. In appendix~\ref{app:Phi-svm},
we show that for any vectors $x^{\pm}\in\reals^{m_{\pm}}$, we have
\[
\Phi(x^{+},x^{-}):=\min_{\nu}\:\sum_{i=1}^{m_{+}}(x_{i}^{+}+\nu)_{+}+\sum_{i=1}^{m_{-}}(x_{i}^{-}-\nu)_{+}=\sum_{i=1}^{\underline{m}}(x_{[i]}^{+}+x_{[i]}^{-})_{+},\]
 with $x_{[j]}$ the $j$-th largest element in a vector $x$. Thus,
the test becomes \[
\lambda>\min_{\mu\ge0}\:\frac{\dsp\sum_{i=1}^{\underline{m}}(2\mu+x_{[i]}^{+}+x_{[i]}^{-})_{+}}{1+(\gamma_{0}/\lambda_{0})\mu}.\]

Setting $\kappa=\lambda_{0}/(\lambda_{0}+\gamma_{0}\mu)$, we obtain
the following formulation for our test: \begin{equation}
{\lambda}>\min_{0\le\kappa\le1}\:\sum_{i=1}^{\underline{m}}((1-\kappa)\frac{2\lambda_{0}}{\gamma_{0}}+\kappa(x_{[i]}^{+}+x_{[i]}^{-}))_{+}=\frac{2\lambda_{0}}{\gamma_{0}}G(\frac{\gamma_{0}}{2\lambda_{0}}\overline{x}),\label{eq:test-svm-kappa}\end{equation}
 where $\overline{x}_{i}:=x_{[i]}^{+}+x_{[i]}^{-}$, $i=1,\ldots,\underline{m}$,
and for $z\in\reals^{m}$, we define \[
G(z):=\min_{0\le\kappa\le1}\:\sum_{i=1}^{m}(1-\kappa+\kappa z_{i})_{+}.\]
 We show in appendix~\ref{app:G-svm} that $G(z)$ admits a closed-form
expression, which can be computed in $O(d\log d)$, where $d$ is
the number of non-zero elements in vector $z$. By construction, the
test removes all the features if we set $\lambda_{0}=\lambda_{{\rm max}}$,
$\gamma_{0}=\gamma_{{\rm max}}$, and when $\lambda>\lambda_{{\rm max}}$.

{\bf Theorem (SAFE-SVM)} {\it Consider the SVM
problem ${\cal P}_{{\rm hi}}(\lambda)$ in~(\ref{eq:pb-svm-primal}).
Denote by $x_{k}$ the $k$-th row of the matrix $[y_{1}z_{1},\ldots,y_{m}z_{m}]$,
and let ${\cal I}_{\pm}:=\{i\::\: y_{i}=\pm1\}$, $m_{\pm}:=|{\cal I}_{\pm}|$,
$\underline{m}:=\min(m_{+},m_{-})$, and $\gamma_{{\rm max}}:=2\underline{m}$.
Let $\lambda_{0}\ge\lambda$ be a value for which the optimal value
$\gamma_{0}\in[0,\gamma_{{\rm max}}]$ of ${\cal P}_{{\rm sq}}(\lambda_{0})$
is known. The following condition allows to remove the $k$-th feature
vector $x_{k}$: \begin{equation}
{\lambda}>\frac{2\lambda_{0}}{\gamma_{0}}\max\left(G(\frac{\gamma_{0}}{2\lambda_{0}}\overline{x}_{k}),G(\frac{\gamma_{0}}{2\lambda_{0}}\underline{x}_{k})\right),\label{eq:test-svm-lambda0}\end{equation}
 where $(\overline{x}_{k})_{i}:=(x_{k})_{[i]}^{+}+(x_{k})_{[i]}^{-}$,
$(\underline{x}_{k})_{i}:=(-x_{k})_{[i]}^{+}+(-x_{k})_{[i]}^{-}$,
$i=1,\ldots,\underline{m}$, and for $z\in\reals^{m}$: \[
G(z)=\min_{z}\:\dsp\frac{1}{1-z}\sum_{i=1}^{p}(z_{i}-z)_{+}~:~z\in\{-\infty,0,(z_{j})_{j\::\: z_{j}<0}\}\]
 A specific choice for $\lambda_{0}$ is $\overline{\lambda}_{{\rm max}}$
given by~(\ref{eq:lambda0-def-svm}), with corresponding optimal
value $\gamma_{0}=\gamma_{{\rm max}}$.} \hfill\BlackBox

\subsection{SAFE for Sparse Logistic Regression}

\label{s:lo}

We now consider the sparse logistic regression problem: \begin{equation}
{\cal P}_{{\rm lo}}(\lambda)~:~~\phi(\lambda):=\min_{w,v}\:\sum_{i=1}^{m}\log\left(1+\exp(-y_{i}(z_{i}^{T}w+v))\right)+\lambda\|w\|_{1},\label{eq:pb-lo-primal}\end{equation}
 with the same notation as in section~\ref{s:hi}. The dual problem takes the form \begin{equation}
{\cal D}_{{\rm lo}}(\lambda)~:~~\phi(\lambda):=\max_{\theta}\:\sum_{i=1}^{m}\left(\theta_{i}\log(-\theta_{i})-(1+\theta_{i})^{T}\log(1+\theta_{i})\right)~:~\begin{array}[t]{l}
-\ones\le\theta\le0,\;\;\theta^{T}y=0,\\
|\theta^{T}x_{k}|\le\lambda,\;\; k=1,\ldots,n.\end{array}\label{eq:pb-lo-dual}\end{equation}

\subsubsection{Test, $\gamma$ given}

Assume that we know a lower bound on the problem, $\gamma\le\phi(\lambda)$.
Since $0\le\phi(\lambda)\le m\log2$, we may assume that $\gamma\in[0,m\log2]$
without loss of generality. We proceed to formulate problem~(\ref{eq:P-def-dual}).
For given $x\in\reals^{m}$, and $\gamma\in\reals$, we have \begin{equation}
\begin{array}{rcl}
P_{{\rm log}}(\gamma,x) & = & \dsp\min_{\mu>0,\:\nu}\:-\gamma\mu+\mu\sum_{i=1}^{m}f_{{\rm log}}\left(\dsp\frac{x_{i}+y_{i}\nu}{\mu}\right),\end{array}\label{eq:P-lo-pb}\end{equation}
 which can be computed in $O(m)$ by two-dimensional search, or by
the dual interior-point method described in appendix. (As mentioned
before, an alternative, resulting in a more conservative test, is
to fix $\nu$, for example $\nu=0$.) Our test to eliminate the $k$-th
feature takes the form \[
\lambda>T_{{\rm log}}(\gamma,x_{k}):=\max(P_{{\rm log}}(\gamma,x_{k}),P_{{\rm log}}(\gamma,-x_{k})).\]
 If $\gamma$ is known, the complexity of running this test through
all the features is $O(nm)$. (In fact, the terms in the objective
function that correspond to zero elements of $x$ are of two types,
involving $f_{{\rm log}}(\pm\nu/\mu)$. This means that the effective
dimension of problem~(\ref{eq:P-lo-pb}) is the cardinality $d$
of vector $x$, which in many applications is much smaller than $m$.)

\subsubsection{Obtaining a dual feasible point}

We can construct dual feasible points based on scaling one obtained
by choice of a primal point (classifier weight) $w_{0}$. This in
turn leads to other possible choices for the bound $\gamma$.

For $w_{0}\in\reals^{n}$ given, we solve the one-dimensional, convex
problem \[
v_{0}:=\arg\min_{b}\:\dsp\sum_{i=1}^{m}f_{{\rm log}}(y_{i}x_{i}^{T}w_{0}+y_{i}b).\]
 This problem can be solved by bisection in $O(m)$ time \cite{boyd_linear}.
At optimum, the derivative of the objective is zero, hence $y^{T}\theta_{0}=0$,
where \[
\theta_{0}(i):=-\frac{1}{1+\exp(y_{i}x_{i}^{T}w_{0}+y_{i}v_{0})},\;\; i=1,\ldots,m.\]
 Now apply the scaling method seen before, and set $\gamma$ by solving
problem~(\ref{eq:gamma-pb}).

\subsubsection{A specific example of a dual point}

A convenient, specific choice in the above construction is to set
$w_{0}=0$. Then, the intercept $v_{0}$ can be explicitly computed,
as $v_{0}=\log(m_{+}/m_{-})$, where $m_{\pm}=|\{i\::\: y_{i}=\pm1\}|$
are the class cardinalities. The corresponding dual point $\theta_{0}$
is \begin{equation}
\theta_{0}(i)=\left\{ \begin{array}{ll}
-\dsp\frac{m_{-}}{m} & (y_{i}=+1)\\[0.1in]
-\dsp\frac{m_{+}}{m} & (y_{i}=-1),\end{array}\right.\;\; i=1,\ldots,m.\label{eq:theta0-logreg}\end{equation}
 The corresponding value of $\lambda_{0}$ is (see \cite{boyd_linear}):
\[
\lambda_{0}:=\|X^{T}\theta_{0}\|_{\infty}=\max_{1\le k\le n}|\theta_{0}^{T}x_{k}|.\]

We now compute $\gamma(\lambda)$ by solving problem~(\ref{eq:gamma-pb}),
which expresses as \begin{equation}
\gamma(\lambda)=\max_{|s|\le\lambda/\lambda_{0}}\: G_{{\rm log}}(s\theta_{0})=\max_{|s|\le\lambda/\lambda_{0}}\:-m_{+}f_{{\rm log}}^{\ast}(-s\frac{m_{-}}{m})-m_{-}f_{{\rm log}}^{\ast}(-s\frac{m_{+}}{m}).\label{eq:theta-logreg}\end{equation}
 The above can be solved analytically: it can be shown that $s=\lambda/\lambda_{0}$
is optimal.

\subsubsection{Solving the bisection problem}

In this section, we are given $c\in\reals^{m}$, $\gamma\in(0,m\log2)$,
and we consider the problem \begin{equation}
\begin{array}{rcl}
F^{\ast}:=\dsp\min_{\mu>0}\: F(\mu) & := & -\gamma\mu+\mu\dsp\sum_{i=1}^{m}f_{{\rm log}}(c(i)/\mu).\end{array}\label{eq:F-logreg}\end{equation}
 Problem~(\ref{eq:F-logreg}) corresponds to the problem~(\ref{eq:P-lo-pb}),
with $\nu$ set to a fixed value, and $c(i)=y_{i}x_{i}$, $i=1,\ldots,m$.
We assume that $c(i)\ne0$ for every $i$, and that $\kappa:=m\log2-\gamma>0$.
Observe that $F^{\ast}\le F_{0}:=\lim_{\mu\rightarrow0^{+}}\: F(\mu)=\ones^{T}c_{+}$,
where $c_{+}$ is the positive part of vector $c$.

To solve this problem via bisection, we initialize the interval of
confidence to be $[0,\mu_{u}]$, with $\mu_{u}$ set as follows. Using
the inequality $\log(1+e^{-x})\ge\log2-(1/2)x_{+}$, which is valid
for every $x$, we obtain that for every $\mu>0$: \[
F(\mu)\ge-\gamma\mu+\mu\sum_{i=1}^{m}\left(\log2-\frac{(c(i))_{+}}{2\mu}\right)=\kappa\mu-\frac{1}{2}\ones^{T}c_{+}.\]
 We can now identify a value $\mu_{u}$ such that for every $\mu\ge\mu_{u}$,
we have $F(\mu)\ge F_{0}$: it suffices to ensure $\kappa\mu-(1/2)\ones^{T}c_{+}\ge F_{0}$,
that is, \[
\mu\ge\mu_{u}:=\frac{(1/2)\ones^{T}c_{+}+F_{0}}{\kappa}=\frac{3}{2}\frac{\ones^{T}c_{+}}{m\log2-\gamma}.\]

\subsubsection{Algorithm summary}

An algorithm to check if a given feature can be removed from a sparse
logistic regression problem works as follows.

{\em Given:} $\lambda$, $k$ ($1\le k\le n$), $f_{{\rm log}}(x)=\log(1+e^{-x})$,
$f_{{\rm log}}^{\ast}(\vartheta)=(-\vartheta)\log(-\vartheta)+(\vartheta+1)\log(\vartheta+1)$. 
\begin{enumerate}
\item Set $\lambda_{0}=\dsp\max_{1\le k\le n}|\theta_{0}^{T}x_{k}|$, where
$\theta_{0}(i)=-m_{-}/m$ ($y_{i}=+1$), $\theta_{0}(i)=-m_{+}/m$
($y_{i}=-1$), $i=1,\ldots,m$. 
\item Set \[
\gamma(\lambda):=-m_{+}f_{{\rm log}}^{\ast}(-\frac{\lambda}{\lambda_{0}}\frac{m_{-}}{m})-m_{-}f_{{\rm log}}^{\ast}(-\frac{\lambda}{\lambda_{0}}\frac{m_{+}}{m}).\]
 
\item Solve via bisection a pair of one-dimensional convex optimization
problems \[
P_{\epsilon}=\dsp\min_{\mu>0}\:-\gamma(\lambda)\mu+\mu\dsp\sum_{i=1}^{m}f_{{\rm log}}(\epsilon y_{i}(x_{k})_{i}/\mu)\;\;(\epsilon=\pm1),\]
 each with initial interval $[0,\mu_{u}]$, with \[
\mu_{u}=\frac{3}{2}\frac{\dsp\sum_{i=1}^{m}(\epsilon y_{i}(x_{k})_{i})_{+}}{m\log2-\gamma}.\]

\item If $\lambda>\max(P_{+},P_{-})$, the $k$-th feature can be safely
removed. 
\end{enumerate}

\section{Numerical results}

\label{s:num} In this section we explore the benefits of SAFE by
running numerical experiments%
\footnote{In our experiments, we have used an Apple Mac Pro 64-bit workstation,
with two $2.26$ GHz Quad-Core Intel Xeon processors, $8$ MB on-chip
shared L3 cache per processor, with $6$ GB SDRAM, operating at $1066$
MHz.%
} with different LASSO solvers. We present two kinds of experiments
to highlight the two main benefits of SAFE. One kind, in our opinion
the most important, shows how memory limitations can be reduced, by
allowing to treat larger data sets. The other focuses on measuring
computational time reduction when using SAFE a priori to the LASSO
solver.

We have used a variety of available algorithms
for solving the LASSO problem. We use acronyms to refer to the following
methods: IPM stands for the Interior-Point Method for LASSO described
in~\cite{boyd_linear}; GLMNET corresponds to the Generalized Linear
Model algorithm described in \cite{FHT:10}; TFOCS corresponds to
Templates for First-Order Conic Solvers described in \cite{becker2010templates};
FISTA and Homotopy stand for the Fast Iterative Shrinkage-Thresholding
Algorithm and homotopy algorithm, described and implemented in \cite{yang2010fast},
respectively. Some methods (like IPM, TFOCS) do not return exact zeros
in the final solution of the LASSO problem and the issue arises in
evaluating the its cardinality. In appendix~\ref{app:thresholding},
we discuss some issue related to the thresholding of the LASSO solution.

In our experiments, we use data sets derived from text classification
sources in \cite{Frank+Asuncion:2010}. We use medical journal abstracts
from PubMed represented in a bag-of-words format, where stop words
have been eliminated and capitalization removed. The dimensions of
the feature matrix $X$ we use from PubMed is $m=1,000,000$ abstracts
and $n=127,025$ features (words). There is a total of $82,209,586$
non-zeros in the feature matrix, with an average of about $645$ non-zeros
per feature (word). We also use data-sets derived from the headlines
of {\em The New York Times,} (NYT) spanning a period of about $20$
years (from 1985 to 2007). The number of headlines in the entire NYT
data-set is $m=3,241,260$ and the number of features (words) is $n=159,943$.
There is a total of $14,083,676$ non-zeros in the feature matrix,
with an average of about $90$ non-zeros per feature.

In some applications such as \cite{GJMEYC:10}, the goal is to learn
a short list of words that are predictive of the appearance of a given
query term (say, ``lung'' or ``china'') in the abstracts of
medical journals or NYT news. The LASSO problem can be used to produce
a summarization of the query term across the many abstracts or headlines
considered. To be manageable by a human reader, the list of predictive
terms should be very short (say at most $100$ terms) with respect
to the size of the dictionary $n$. To produce such a short list,
we solve the LASSO problem (\ref{eq:LASSO}) with different penalty
parameters $\lambda$, and choose the appropriate penalty $\lambda$
that would generate enough non-zeros in the LASSO solution (around
$100$ non-zeros in our case).

\subsection{SAFE for reducing memory limit problems}
\begin{figure}
\begin{centering}
\includegraphics[width=12cm]{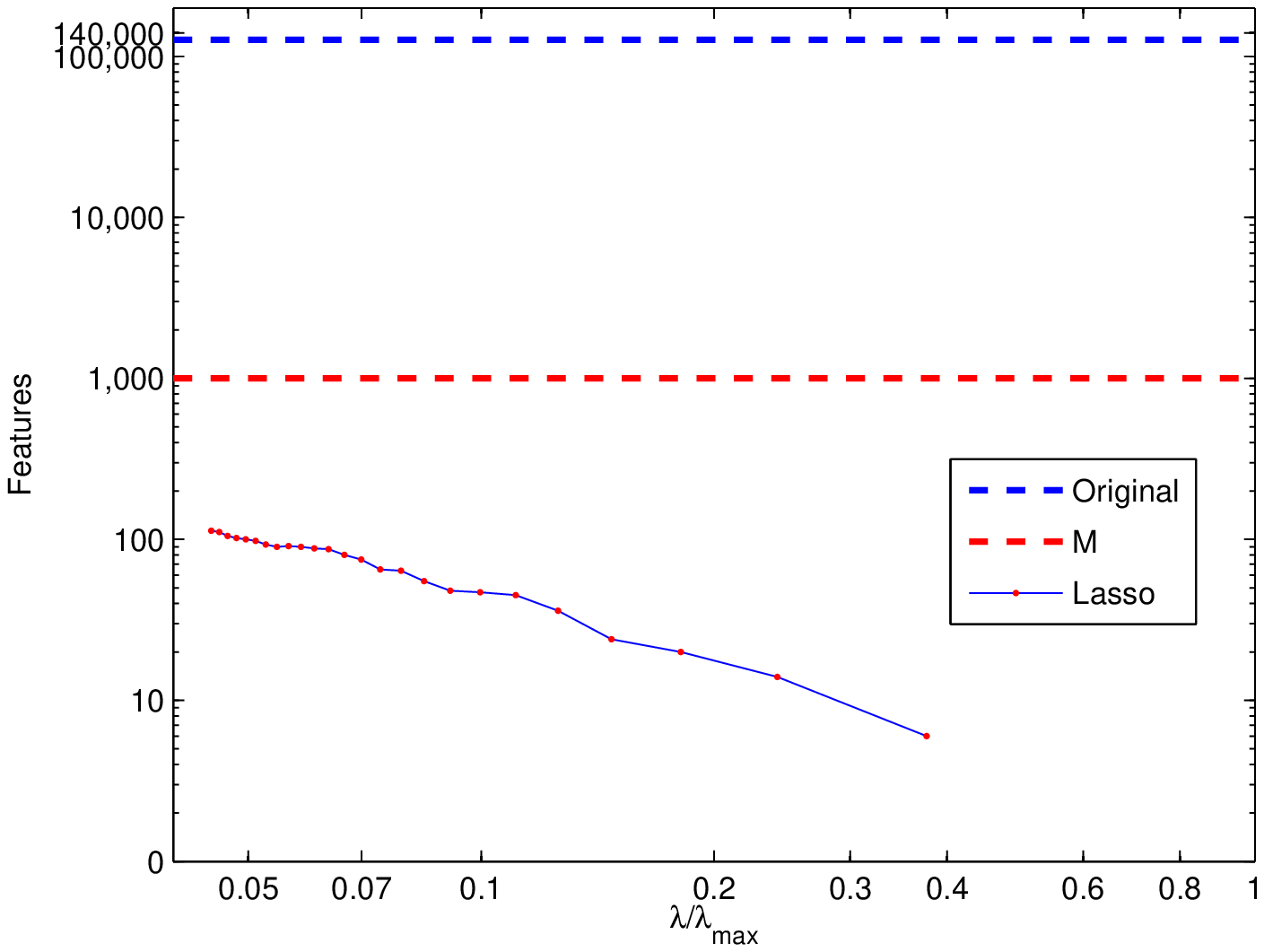} 
\par\end{centering}

\caption{\label{fig:mem_lim_pubmed}A LASSO problem solved for the PubMed data-set
and $\lambda=0.04\lambda_{max}$ using a sequence of $25$ smaller
size problems. Each LASSO problem in the sequence has a number of
features $L_{F}$ that satisfies the memory limit $M=1,000$, i.e
$L_{F}\leq1,000$. }

\end{figure}

We experiment with PubMed data-set which is too large to be
loaded into memory, and thus not amenable to current LASSO solvers. As described before, we are interested
in solving the LASSO problem for a regularization parameter that would
result in about $100$ non-zeros in the solution. We implement algorithm
\ref{alg:SAFE-for-memory} with a memory limit $M=1,000$ features,
where we have observed that for the PubMed data loading more than
$1,000$ features causes memory problems in the machine and platform
we are using. The memory limit is approximately two orders of magnitudes
less than the original number of features $n$, i.e. $M\approx0.01n$.
Using algorithm \ref{alg:SAFE-for-memory}, we were able to solved
the LASSO problem for $\lambda=0.04\lambda_{max}$ using a sequence
of $25$ LASSO problem with each problem having a number of features
less than $M=1,000$. Figure \ref{fig:mem_lim_pubmed} shows the simulation
result for the PubMed data-set.%

\subsection{SAFE for LASSO run-time reduction}
\begin{figure}
\subfigure[]{\label{subfig:perc_savings}\includegraphics[width=8cm]{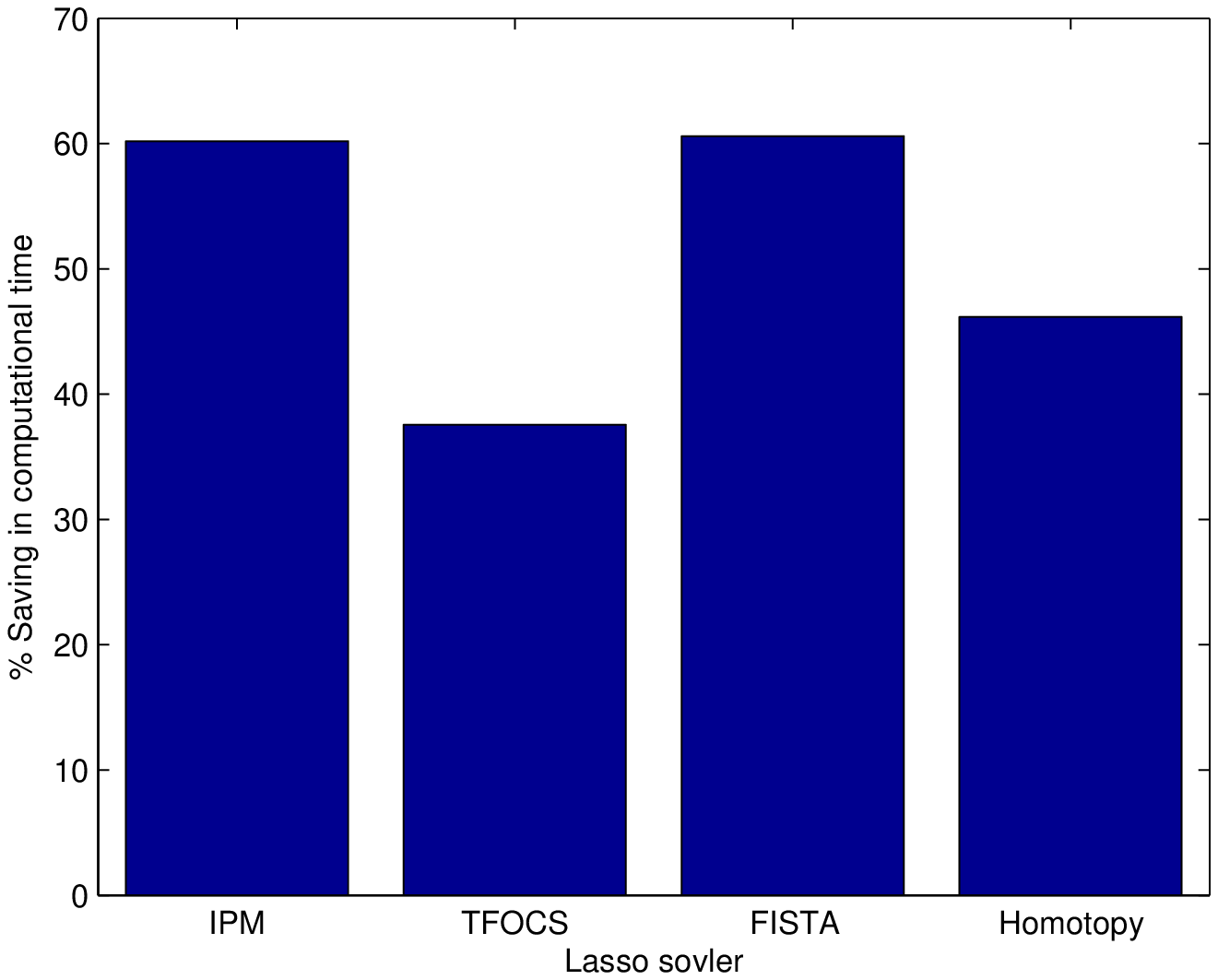}}\hfill{}\subfigure[]{\label{subfig:nyt_soln}\includegraphics[width=8cm]{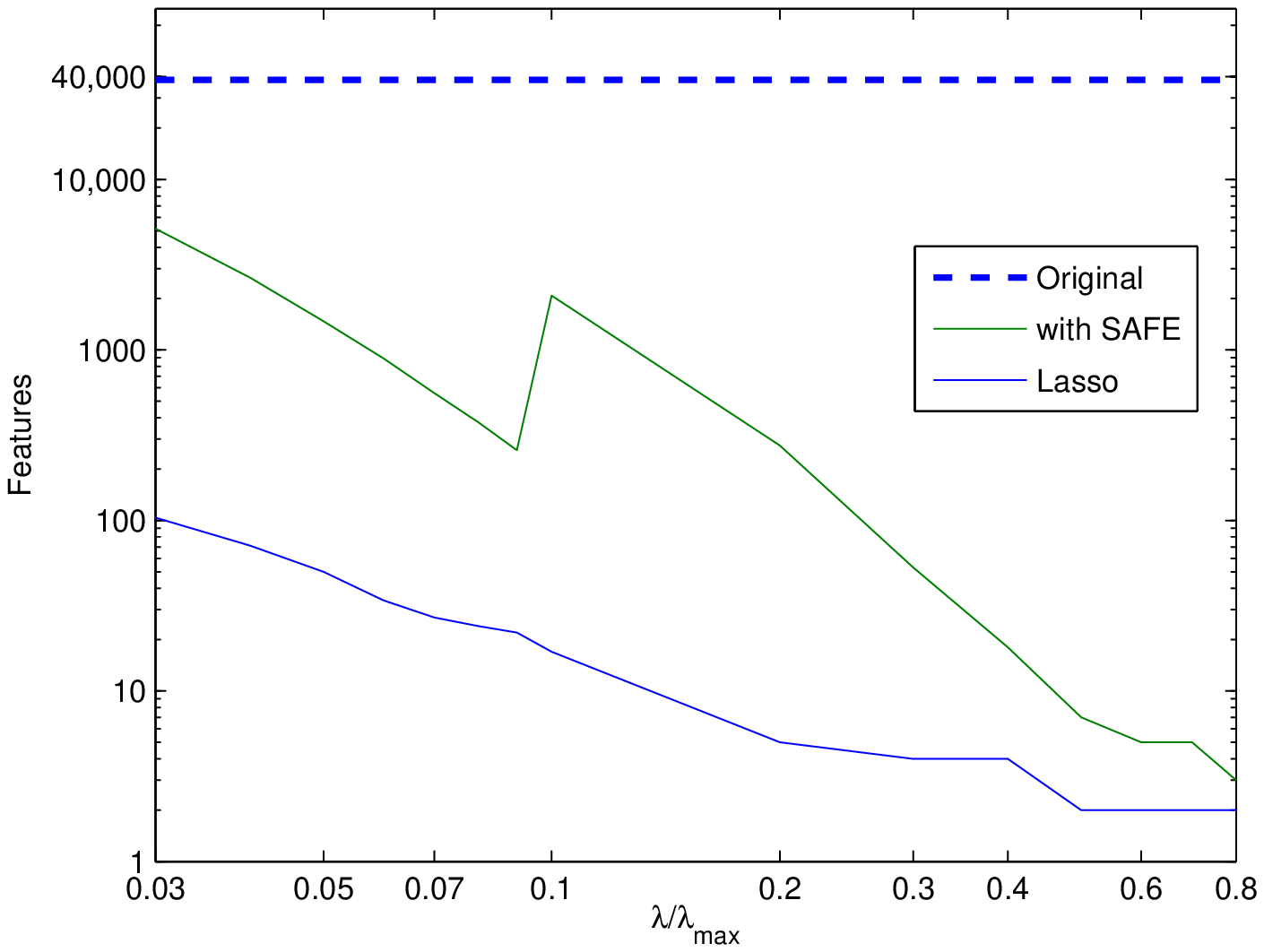}}
\caption{(a) Computational time savings. (b) Lasso solution for the sequence
of problem between $0.03\lambda_{max}$ and $\lambda_{max}$. The
green line shows the number of features we used to solve the LASSO
problem after using algoirthm \ref{alg:Recursive-SAFE-for-1}.\label{fig:savings}}

\end{figure}

We have used a portion of the NYT data-set corresponding to all headlines
in year $1985,$ the corresponding feature matrix has dimensions $n=38,377$
features and $m=192,182$ headlines, with an average of $21$ non-zero
per feature. We solved the plain LASSO problem and the LASSO problem
with SAFE as outlined in algoirthm \ref{alg:Recursive-SAFE-for-1}
for a sequence of $\lambda$ logarithmically distributed between $0.03\lambda_{max}$
and $\lambda_{max}$. We have used four LASSO solvers, IPM, TFOCS,
FISTA and Homotopy to solve the LASSO problem. Figure \ref{subfig:perc_savings}shows
the computational time saving when using SAFE. Figure \ref{subfig:nyt_soln}
shows the number of features we used to solve the LASSO problem when
using SAFE, and the number of non-zeros in the solution. We realize
that when using algorithm \ref{alg:Recursive-SAFE-for-1} we solve
problems with a number of features at most $10,000$ instead of $n=38,377$
features, this reduction has a direct impact on the solving time of
the LASSO problem as demonstrated in figure \ref{subfig:perc_savings}.

\subsection{SAFE for LASSO with intercept problem} 
\begin{figure}
\begin{centering}
\includegraphics[width=12cm]{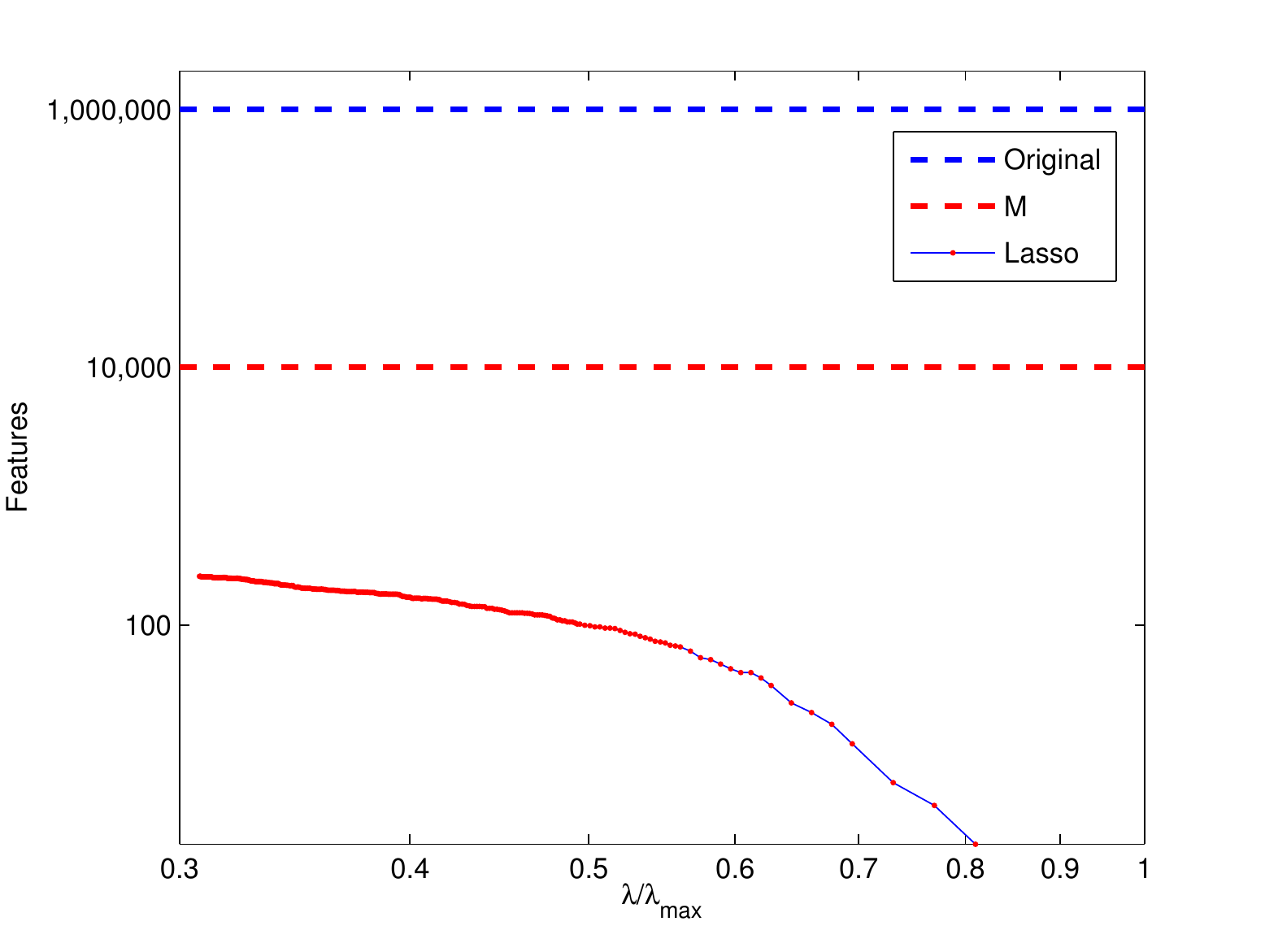} 
\par\end{centering}

\caption{\label{fig:mem_lim_glmnet}A LASSO problem with intercept solved for
randomly generated data-set and $\lambda=0.33\lambda_{max}$ using
a sequence of $352$ smaller size problems. Each LASSO problem in
the sequence has a number of features $L_{F}$ that satisfies the
memory limit $M=10,000$, i.e $L_{F}\leq1000$. }

\end{figure}

We return to the LASSO with intercept problem discussed in section~\ref{ss:intercept-lasso}.
We generate a feature matrix $X\in\reals^{m\times n}$ with $m=500$,
$n=10^{6}$. The entries of $X$ has a ${\cal N}(0,1)$ normal distributed
and sparsity density $d=0.1$.  We also generate a vector of coefficients
$\omega\in\reals^{n}$ with $50$ non-zero entries. The response $y$
is generated by setting $y=X\omega+0.01\eta$, where $\eta$ is a
vector in $\reals^{m}$ with ${\cal N}(0,1)$ distribution. We use
GLMNET implemented in \verb"R" to solve the LASSO problem with intercept.
The generated data, $X$ and $y$ can be loaded into \verb"R" , yet
memory problems occur when we try to solve the LASSO problem. We use
algorithm \ref{alg:SAFE-for-memory} with memory limit $M=10,000$
features and $\lambda=0.33\lambda_{max}$. Figure \ref{fig:mem_lim_glmnet}
shows the number of non-zeros in the solution of the $352$ sequence
of problems used to obtain the solution at $\lambda=0.33\lambda_{max}$.

\newpage
\appendix

\section{Expression of $P(\gamma,x_k)$ (LASSO)}
\label{app:p-gamma}
We can express problem (\ref{eq:testproblem}) in dual form as a convex
optimization problem with two scalar variables, $\mu_{1}$ and $\mu_{2}$:

\begin{eqnarray*}
P(\gamma,x_{k}) & = & \;\min_{\mu_{1},\mu_{2}\geq0}\max_{\theta}\, x_{k}^{T}\theta+\mu_{1}\left(G(\theta)-\gamma\right)+\mu_{2}g^{T}\left(\theta-\theta_{0}^{\star}\right)\\
 & = & \;\min_{\mu_{1},\mu_{2}\geq0}-\mu_{1}\gamma-\mu_{2}g^{T}\theta_{0}^{\star}+\max_{\theta}\, x_{k}^{T}\theta+\mu_{1}G(\theta)+\mu_{2}g^{T}\theta\\
 & = & \;\min_{\mu_{1},\mu_{2}\geq0}-\mu_{1}\gamma-\mu_{2}g^{T}\theta_{0}^{\star}+\mu_{1}\max_{\theta}\,\left(\frac{x_{k}^{T}-\mu_{1}y^{T}+\mu_{2}g^{T}}{\mu_{1}}\theta-\frac{1}{2}\left\Vert \theta\right\Vert _{2}^{2}\right)\end{eqnarray*}

We obtain:\begin{eqnarray}
P(\gamma,x_{k}) & = & \min_{\mu_{1},\mu_{2}\geq0}L(\mu_{1},\mu_{2})\label{eq:safe3test_1}\end{eqnarray}
with \begin{equation}
L(\mu_{1},\mu_{2})=-x_{k}^{T}y+\frac{\mu_{1}}{2}D^{2}+\frac{1}{2\mu_{1}}\left\Vert x_{k}\right\Vert _{2}^{2}+\frac{\mu_{2}^{2}}{2\mu_{1}}\left\Vert g\right\Vert _{2}^{2}+\frac{\mu_{2}}{\mu_{1}}x_{k}^{T}g-\mu_{2}\left\Vert g\right\Vert _{2}^{2},\label{eq:objec_of_test}\end{equation}
and $D:=\left(\left\Vert y\right\Vert _{2}^{2}-2\gamma\right)^{1/2}$.

To solve (\ref{eq:safe3test_1}), we take the derivative of (\ref{eq:objec_of_test})
w.r.t $\mu_{2}$ and set it to zero:

\[
\mu_{2}\left\Vert g\right\Vert _{2}^{2}+x_{k}^{T}g-\mu_{1}\left\Vert g\right\Vert _{2}^{2}=0.\]

This implies that $\mu_{2}=\text{max}(0,\mu_{1}-\frac{x_{k}^{T}g}{\left\Vert g\right\Vert _{2}^{2}})$.
When $\mu_{1}\leq\frac{x_{k}^{T}g}{\left\Vert g\right\Vert _{2}^{2}}$,
we have $\mu_{2}=0$, $\mu_{1}=\frac{\left\Vert x_{k}\right\Vert _{2}}{D}$
and $P(\gamma,x_{k})$ takes the value:

\[
P(\gamma,x_{k})=-y^{T}x_{k}+\left\Vert x_{k}\right\Vert _{2}D.\]

On the other hand, when $\mu_{1}\geq\frac{x_{k}^{T}g}{\left\Vert g\right\Vert _{2}^{2}}$,
we take the derivative of (\ref{eq:objec_of_test}) w.r.t $\mu_{1}$
and set it to zero:

\[
\tilde{D}^{2}\mu_{1}^{2}=\Psi_{k}^{2},\]
with $\Psi_{k}=\left(\left\Vert x_{k}\right\Vert _{2}^{2}-\frac{\left(x_{k}^{T}g\right)^{2}}{\left\Vert g\right\Vert _{2}^{2}}\right)^{1/2}$
and $\tilde{D}=\left(D^{2}-\left\Vert g\right\Vert _{2}^{2}\right)^{1/2}$.
Substituting $\mu_{1}$ and $\mu_{2}$ in (\ref{eq:safe3test_1}),
$P(\gamma,x_{k})$ takes the value:

\[
P(\gamma,x_{k})=\theta_{0}^{\star T}x_{k}+\Psi_{k}\tilde{D}.\]

\section{Expression of $P(\gamma,x)$, general case}

We show that the quantity $P(\gamma,x)$ defined in~(\ref{eq:P-def})
can be expressed in dual form~(\ref{eq:P-def-dual}). This is a simple
consequence of duality: \[
\begin{array}{rcl}
P(\gamma,x) & = & \dsp\max_{\theta}\:\theta^{T}x~:~G(\theta)\ge\gamma,\;\;\theta^{T}b=0\\
 & = & \dsp\max_{\theta}\:\min_{\mu>0,\:\nu}\:\theta^{T}x+\mu(G(\theta)-\gamma)-\nu\theta^{T}b\\
 & = & \dsp\min_{\mu>0,\:\nu}\:\dsp\max_{\theta}\:\theta^{T}x+\mu(-y^{T}\theta-\sum_{i=1}^{m}f^{\ast}(\theta(i))-\gamma)-\nu\theta^{T}b\\
 & = & \dsp\min_{\mu>0,\:\nu}\:-\gamma\mu+\dsp\max_{\theta}\:\theta^{T}(x-\mu y-\nu z)-\mu\sum_{i=1}^{m}f^{\ast}(\theta(i))\\
 & = & \dsp\min_{\mu>0,\:\nu}\:-\gamma\mu+\mu\left(\dsp\max_{\theta}\:\frac{1}{\mu}\theta^{T}(x-\mu y-\nu z)-\sum_{i=1}^{m}f^{\ast}(\theta(i))\right)\\
 & = & \dsp\min_{\mu>0,\:\nu}\:-\gamma\mu+\mu\sum_{i=1}^{m}f\left(\frac{x_{i}-\mu y(i)-\nu b_{i}}{\mu}\right).\end{array}\]

\section{SAFE test for SVM}

\label{app:min-poly-2d} In this section, we examine various optimization
problems involving polyhedral functions in one or two variables, which
arise in section~\ref{ss:test-gamma-svm} for the computation of
$P_{{\rm hi}}(\gamma,x)$ as well as in the SAFE-SVM theorem of section~\ref{ss:safe-thm-svm}.

\subsection{Computing $P_{{\rm hi}}(\gamma,x)$}

\label{app:P-gamma-svm} We first focus on the specific problem of
computing the quantity defined in~(\ref{eq:P-hi-pb}). To simplify
notation, we will consider the problem of computing $P_{{\rm hi}}(\gamma,-x)$,
that is: \begin{equation}
P_{{\rm hi}}(\gamma,-x)=\dsp\min_{\mu\ge0,\:\nu}\:-\gamma\mu+\sum_{i=1}^{m}(\mu+\nu y_{i}+x_{i})_{+},\label{eq:app-P-hi}\end{equation}
 where $y\in\{-1,1\}^{m}$, $x\in\reals^{m}$ and $\gamma$ are given,
with $0\le\gamma\le\gamma_{0}:=2\min(m_{+},m_{-})$. Here, ${\cal I}_{\pm}:=\{i\::\: y_{i}=\pm1\}$,
and $x^{+}=(x_{i})_{i\in{\cal I}_{+}}$, $x^{-}=(x_{i})_{i\in{\cal I}_{-}}$,
$m_{\pm}=|{\cal I}_{\pm}|$, and $\underline{m}=\min(m_{+},m_{-})$.
Without loss of generality, we assume that both $x^{+},x^{-}$ are
both sorted in descending order: $x_{1}^{\pm}\ge\ldots\ge x_{m_{\pm}}^{\pm}$.

Using $\alpha=\mu+\nu$, $\beta=\mu-\nu$, we have \begin{equation}
\begin{array}{rcl}
P_{{\rm hi}}(\gamma,-x) & = & \dsp\min_{\alpha+\beta\ge0}\:-\dsp\frac{\gamma}{2}(\alpha+\beta)+\sum_{i=1}^{m_{+}}(x_{i}^{+}+\alpha)_{+}+\sum_{i=1}^{m_{-}}(x_{i}^{-}+\beta)_{+}\\
 & = & \dsp\min_{\alpha,\:\beta}\:\max_{t\ge0}\:-\dsp\frac{\gamma}{2}(\alpha+\beta)+\sum_{i=1}^{m_{+}}(x_{i}^{+}+\alpha)_{+}+\sum_{i=1}^{m_{-}}(x_{i}^{-}+\beta)_{+}-t(\alpha+\beta)\\
 & = & \dsp\max_{t\ge0}\:\dsp\min_{\alpha,\:\beta}\:-(\dsp\frac{\gamma}{2}+t)(\alpha+\beta)+\sum_{i=1}^{m_{+}}(x_{i}^{+}+\alpha)_{+}+\sum_{i=1}^{m_{-}}(x_{i}^{-}+\beta)_{+}\\
 & = & \dsp\max_{t\ge0}\: F(\dsp\frac{\gamma}{2}+t,x^{+})+F(\dsp\frac{\gamma}{2}+t,x^{-}),\end{array}\label{eq:P-hi-via-F}\end{equation}
 where, for $h\in\reals$ and $x\in\reals^{p}$, $x_{1}\ge\ldots\ge x_{p}$,
we set \begin{equation}
F(h,x):=\min_{z}\:-hz+\sum_{i=1}^{p}(z+x_{i})_{+},\label{eq:P-hi-F-xh}\end{equation}


\paragraph{Expression of the function $F$.}

If $h>p$, then with $z\rightarrow+\infty$ we obtain $F(h,x)=-\infty$.
Similarly, if $h<0$, then $z\rightarrow-\infty$ yields $F(h,x)=-\infty$.
When $0\le h\le p$, we proceed by expressing $F$ in dual form: \[
F(h,x)=\max_{u}\: u^{T}x~:~0\le u\le\ones,\;\; u^{T}\ones=h.\]

If $h=p$, then the only feasible point is $u=\ones$, so that $F(p,x)=\ones^{T}x$.
If $0\le h<1$, choosing $u_{1}=h$, $u_{2}=\ldots=u_{p}=0$, we obtain
the lower bound $F(h,x)\ge hx_{1}$, which is attained with $z=-x_{1}$.

Assume now that $1\le h<p$. Let $h=q+r$, with $q=\lfloor h\rfloor$
the integer part of $h$, and $0\le r<1$. Choosing $u_{1}=\ldots=u_{q}=1$,
$u_{q+1}=r$, we obtain the lower bound \[
F(h,x)\ge\sum_{j=1}^{q}x_{j}+rx_{q+1},\]
 which is attained by choosing $z=-x_{q+1}$ in the expression~(\ref{eq:P-hi-F-xh}).

To summarize: \begin{equation}
F(h,x)=\left\{ \begin{array}{ll}
hx_{1} & \mbox{if }0\le h<1,\\
\dsp\sum_{j=1}^{\lfloor h\rfloor}x_{j}+(h-\lfloor h\rfloor)x_{\lfloor h\rfloor+1} & \mbox{if }1\le h<p,\\
\dsp\sum_{j=1}^{p}x_{j} & \mbox{if }h=p,\\
-\infty & \mbox{otherwise.}\end{array}\right.\label{eq:P-hi-F-xh-soln}\end{equation}
 A more compact expression, valid for $0\le h\le p$ if we set $x_{p+1}=x_{p}$
and assume that a sum over an empty index sets is zero, is \[
F(h,x)=\dsp\sum_{j=1}^{\lfloor h\rfloor}x_{j}+(h-\lfloor h\rfloor)x_{\lfloor h\rfloor+1},\;\;0\le h\le p.\]
 Note that $F(\cdot,x)$ is the piece-wise linear function that interpolates
the sum of the $h$ largest elements of $x$ at the integer break
points $h=0,\ldots,p$.

\paragraph{Expression of $P_{{\rm hi}}(\gamma,-x)$.}

We start with the expression found in~(\ref{eq:P-hi-via-F}): \[
P_{{\rm hi}}(\gamma,-x)=\dsp\max_{t\ge0}\: F(\dsp\frac{\gamma}{2}+t,x^{+})+F(\dsp\frac{\gamma}{2}+t,x^{-}).\]
 Since the domain of $F(\cdot,x^{+})+F(\cdot,x^{-})$ is $[0,\underline{m}]$,
and with $0\le\gamma/2\le\gamma_{0}/2=\underline{m}$, we get \[
P_{{\rm hi}}(\gamma,-x)=\dsp\max_{\gamma/2\le h\le\underline{m}}\: G(h,x^{+},x^{-}):=F(h,x^{+})+F(h,x^{-}).\]
 Since $F(\cdot,x)$ with $x\in\reals^{p}$ is a piece-wise linear
function with break points at $0,\ldots,p$, a maximizer of $G(\cdot,x^{+},x^{-})$
over $[\gamma/2,\underline{m}]$ lies in $\{\gamma/2,\lfloor\gamma/2\rfloor+1,\ldots,\underline{m}\}$.
Thus, \[
P_{{\rm hi}}(\gamma,-x)=\dsp\max\left(G(\frac{\gamma}{2},x^{+},x^{-}),\max_{h\in\{\lfloor\gamma/2\rfloor+1,\ldots,\underline{m}\}}\: G(h,x^{+},x^{-})\right).\]

Let us examine the second term, and introduce the notation $\bar{x}_{j}:=x_{j}^{+}+x_{j}^{-}$,
$j=1,\ldots,\underline{m}$: \begin{eqnarray*}
\max_{h\in\{\lfloor\gamma/2\rfloor+1,\ldots,\underline{m}\}}\: G(h,x^{+},x^{-}) & = & \max_{h\in\{\lfloor\gamma/2\rfloor+1,\ldots,\underline{m}\}}\:\sum_{j=1}^{h}(x_{j}^{+}+x_{j}^{-})\\
 & = & \sum_{j=1}^{\lfloor\gamma/2\rfloor+1}\bar{x}_{j}+\sum_{j=\lfloor\gamma/2\rfloor+2}^{\underline{m}}(\bar{x}_{j})_{+},\end{eqnarray*}
 with the convention that sums over empty index sets are zero. Since
\[
G(\frac{\gamma}{2},x^{+},x^{-})=\dsp\sum_{j=1}^{\lfloor\gamma/2\rfloor}\bar{x}_{j}+(\frac{\gamma}{2}-\lfloor\frac{\gamma}{2}\rfloor)\bar{x}_{\lfloor\gamma/2\rfloor+1},\]
 we obtain \[
P_{{\rm hi}}(\gamma,-x)=\dsp\sum_{j=1}^{\lfloor\gamma/2\rfloor}\bar{x}_{j}+\dsp\max\left((\frac{\gamma}{2}-\lfloor\frac{\gamma}{2}\rfloor)\bar{x}_{\lfloor\gamma/2\rfloor+1},\bar{x}_{\lfloor\gamma/2\rfloor+1}+\sum_{j=\lfloor\gamma/2\rfloor+2}^{\underline{m}}(\bar{x}_{j})_{+}\right).\]
 An equivalent expression is: \[
\begin{array}{rcl}
P_{{\rm hi}}(\gamma,-x) & = & \dsp\sum_{j=1}^{\lfloor\gamma/2\rfloor}\bar{x}_{j}-(\frac{\gamma}{2}-\lfloor\frac{\gamma}{2}\rfloor)(-\bar{x}_{\lfloor\gamma/2\rfloor+1})_{+}+\sum_{j=\lfloor\gamma/2\rfloor+1}^{\underline{m}}(\bar{x}_{j})_{+},\;\;0\le\gamma\le2\underline{m},\\
 &  & \bar{x}_{j}:=x_{j}^{+}+x_{j}^{-},\;\; j=1,\ldots,\underline{m}.\end{array}\]
 The function $P_{{\rm hi}}(\cdot,-x)$ linearly interpolates the
values obtained for $\gamma=2q$ with $q$ integer in $\{0,\ldots,\underline{m}\}$:
\[
P_{{\rm hi}}(2q,-x)=\dsp\sum_{j=1}^{q}\bar{x}_{j}+\sum_{j=q+1}^{\underline{m}}(\bar{x}_{j})_{+}.\]

\subsection{Computing $\Phi(x^{+},x^{-})$}

\label{app:Phi-svm} Let us consider the problem of computing \[
\Phi(x^{+},x^{-}):=\min_{\nu}\:\sum_{i=1}^{m_{+}}(x_{i}^{+}+\nu)_{+}+\sum_{i=1}^{m_{-}}(x_{i}^{-}-\nu)_{+},\]
 with $x^{\pm}\in\reals^{m_{\pm}}$, $x_{1}^{\pm}\ge\ldots\ge x_{m_{\pm}}^{\pm}$,
given. We can express $\Phi(x^{+},x^{-})$ in terms of the function
$F$ defined in~(\ref{eq:P-hi-F-xh}): \[
\begin{array}{rcl}
\Phi(x^{+},x^{-}) & = & \dsp\min_{\nu_{+},\nu_{-}}\:\sum_{i\in{\cal I}_{+}}(x_{i}^{+}+\nu^{+})_{+}+\sum_{i\in{\cal I}_{-}}(x_{i}^{-}-\nu^{-})_{+}~:~\nu^{+}=\nu^{-}\\
 & = & \dsp\max_{h}\:\dsp\min_{\nu^{+},\nu^{-}}\:-h(\nu^{+}-\nu^{-})+\sum_{i\in{\cal I}_{+}}(x_{i}^{+}+\nu^{+})_{+}+\sum_{i\in{\cal I}_{-}}(x_{i}^{-}-\nu^{-})_{+}\\
 & = & \dsp\max_{h}\:\dsp\min_{\nu^{+},\nu^{-}}\:-h\nu^{+}+\sum_{i\in{\cal I}_{+}}(x_{i}^{+}+\nu^{+})_{+}+h\nu^{-}+\sum_{i\in{\cal I}_{-}}(x_{i}^{-}-\nu^{-})_{+}\\
 & = & \dsp\max_{h}\:\left(\dsp\min_{\nu}\:-h\nu+\sum_{i\in{\cal I}_{+}}(x_{i}^{+}+\nu)_{+}\right)+\left(\dsp\min_{\nu}\:-h\nu+\sum_{i\in{\cal I}_{-}}(x_{i}^{-}+\nu)_{+}\right)\;\;(\nu_{+}=-\nu_{-}=\nu)\\
 & = & \dsp\max_{h}\: F(h,x^{+})+F(h,x^{-})\\
 & = & \dsp\max_{0\le h\le\underline{m}}\: F(h,x^{+})+F(h,x^{-})\\
 & = & \max(A,B,C),\end{array}\]
 where $F$ is defined in~(\ref{eq:P-hi-F-xh}), and \[
A=\dsp\max_{0\le h<1}\: F(h,x^{+})+F(h,x^{-}),\;\; B:=\dsp\max_{1\le h<\underline{m}}\: F(h,x^{+})+F(h,x^{-})),\;\; C=F(\underline{m},x^{+})+F(\underline{m},x^{-}).\]

We have \[
A:=\dsp\max_{0\le h<1}\: F(h,x^{+})+F(h,x^{-})=\dsp\max_{0\le h<1}\: h(x_{1}^{+}+x_{1}^{-})=(x_{1}^{+}+x_{1}^{-})_{+}.\]
 Next: \begin{eqnarray*}
B & = & \max_{1\leq h<\underline{m}}\: F(h,x^{+})+F(h,x^{-})\\
 & = & \max_{q\in\{1,\ldots ,\underline{m}-1\},r\in[0,1[}\:\sum_{i=1}^{q}(x_{i}^{+}+x_{i}^{-})+r(x_{q+1}^{+}+x_{q+1}^{-})\\
 & = & \max_{q\in\{1,\ldots,\underline{m}-1\}}\sum_{i=1}^{q}(x_{i}^{+}+x_{i}^{-})+(x_{q+1}^{+}+x_{q+1}^{-})_{+}\\
 & = & (x_{1}^{+}+x_{1}^{-})+\sum_{i=2}^{\underline{m}}(x_{i}^{+}+x_{i}^{-})_{+}.\end{eqnarray*}
 Observe that \[
B\geq C=\sum_{i=1}^{\underline{m}}(x_{i}^{+}+x_{i}^{-}).\]
 Moreover, if $(x_{1}^{+}+x_{1}^{-})\geq0$, then $B=\sum_{i=1}^{\underline{m}}(x_{i}^{+}+x_{i}^{-})_{+}\geq A$.
On the other hand, if $x_{1}^{+}+x_{1}^{-}\leq0$, then $x_{i}^{+}+x_{i}^{-}\le0$
for $2\le j\le\underline{m}$, and $A=\sum_{i=1}^{\underline{m}}(x_{i}^{+}+x_{i}^{-})_{+}\geq x_{1}^{+}+x_{1}^{-}=B$.
In all cases, \[
\Phi(x^{+},x^{-})=\max(A,B,C)=\sum_{i=1}^{\underline{m}}(x_{i}^{+}+x_{i}^{-})_{+}.\]

\subsection{SAFE-SVM test}

\label{app:G-svm} Now we consider the problem that arises in the
SAFE-SVM test~(\ref{eq:test-svm-kappa}): \[
G(z):=\min_{0\le\kappa\le1}\:\sum_{i=1}^{p}(1-\kappa+\kappa z_{i})_{+},\]
 where $z\in\reals^{p}$ is given. (The SAFE-SVM condition~(\ref{eq:test-svm-kappa})
involves $z_{i}=\gamma_{0}/(2\lambda_{0})(x_{[i]}^{+}+x_{[i]}^{-})$,
$i=1,\ldots,p:=\underline{m}$.) We develop an algorithm to compute
the quantity $G(z)$, the complexity of which grows as $O(d\log d)$,
where $d$ is (less than) the number of non-zero elements in $z$.

Define ${\cal I}_{\pm}=\{i\::\:\pm z_{i}>0\}$, $k:=|{\cal I}_{+}|$,
$h:=|{\cal I}_{-}|$, $l={\cal I}_{0}$, $l:=|{\cal I}_{0}|$.

If $k=0$, ${\cal I}_{+}$ is empty, and $\kappa=1$ achieves the
lower bound of $0$ for $G(z)$. If $k>0$ and $h=0$, that is, $k+l=p$,
then ${\cal I}_{-}$ is empty, and an optimal $\kappa$ is attained
in $\{0,1\}$. In both cases (${\cal I}_{+}$ or ${\cal I}_{-}$ empty),
we can write \[
G(z)=\min_{\kappa\in\{0,1\}}\:\sum_{i=1}^{p}(1-\kappa+\kappa z_{i})_{+}=\min\left(p,S_{+}\right),\;\; S_{+}:=\sum_{i\in{\cal I}_{+}}z_{i},\]
 with the convention that a sum over an empty index set is zero.

Next we proceed with the assumption that $k\ne0$ and $h\ne0$. Let
us re-order the elements of ${\cal I}_{-}$ in decreasing fashion,
so that $z_{i}>0=z_{k+1}=\ldots=z_{k+l}>z_{k+l+1}\ge\ldots\ge z_{p}$,
for every $i\in{\cal I}_{+}$. (The case when ${\cal I}_{0}$ is empty
is handled simply by setting $l=0$ in our formula.) We have \[
G(z)=k+l+\min_{0\le\kappa\le1}\:\left\{ \kappa\alpha+\sum_{i=k+l+1}^{p}(1-\kappa+\kappa z_{i})_{+}\right\} ,\]
 where, $\alpha:=S_{+}-k-l$. The minimum in the above is attained
at $\kappa=0,1$ or one of the break points $1/(1-z_{j})\in(0,1)$,
where $j\in\{k+l+1,\ldots,p\}$. At $\kappa=0,1$, the objective function
of the original problem takes the values $S_{+},p$, respectively.
The value of the same objective function at the break point $\kappa=1/(1-z_{j})$,
$j=k+l+1,\ldots,p$, is $k+l+G_{j}(z)$, where \begin{eqnarray*}
G_{j}(z) & := & \frac{\alpha}{1-z_{j}}+\dsp\sum_{i=k+l+1}^{p}\left(\frac{z_{i}-z_{j}}{1-z_{j}}\right)_{+}\\
 & = & \frac{\alpha}{1-z_{j}}+\dsp\frac{1}{1-z_{j}}\sum_{i=k+l+1}^{j-1}(z_{i}-z_{j})\\
 & = & \dsp\frac{1}{1-z_{j}}\left(\alpha-(j-k-l-1)z_{j}+\dsp\sum_{i=k+l+1}^{j-1}z_{i}\right)\\
 & = & \dsp\frac{1}{1-z_{j}}\left(S_{+}-(j-1)z_{j}-(k+l)(1-z_{j})+\dsp\sum_{i=k+l+1}^{j-1}z_{i}\right)\\
 & = & -(k+l)+\dsp\frac{1}{1-z_{j}}\left(\dsp\sum_{i=1}^{j-1}z_{i}-(j-1)z_{j}\right).\end{eqnarray*}
 This allows us to write \[
G(z)=\min\left(p,\sum_{i=1}^{k}z_{i},\min_{j\in\{k+l+1,\ldots,p\}}\:\dsp\frac{1}{1-z_{j}}\left(\dsp\sum_{i=1}^{j-1}z_{i}-(j-1)z_{j}\right)\right).\]
 The expression is valid when $k+l=p$ ($h=0$, ${\cal I}_{-}$ is
empty), $l=0$ (${\cal I}_{0}$ is empty), or $k=0$ (${\cal I}_{+}$
is empty) with the convention that the sum (resp.\ minimum) over
an empty index set is $0$ (resp.\ $+\infty$).

We can summarize the result with the compact formula: \[
G(z)=\min_{z}\:\dsp\frac{1}{1-z}\sum_{i=1}^{p}(z_{i}-z)_{+}~:~z\in\{-\infty,0,(z_{j})_{j\::\: z_{j}<0}\}.\]

Let us detail an algorithm for computing $G(z)$. Assume $h>0$. The
quantity \[
\underline{G}(z):=\min_{k+l+1\le j\le p}\:(G_{j}(z))\]
 can be evaluated in less than $O(h)$, via the following recursion:
\begin{equation}
\begin{array}{rcl}
G_{j+1}(z) & = & \dsp\frac{1-z_{j}}{1-z_{j+1}}G_{j}(z)-j\frac{z_{j+1}-z_{j}}{1-z_{j+1}}\\
\underline{G}_{j+1}(z) & = & \min(\underline{G}_{j}(z),G_{j+1}(z))\end{array},\;\; j=k+l+1,\ldots,p,\label{eq:rec-underlineG-svm-app}\end{equation}
 with initial values \[
G_{k+l+1}(z)=\underline{G}_{k+l+1}(z)=\dsp\frac{1}{1-z_{k+l+1}}\left(\dsp\sum_{i=1}^{k+l}z_{i}-(k+l)z_{k+l+1}\right).\]
 On exit, $\underline{G}(z)=\underline{G}_{p}$.

Our algorithm is as follows.

\bigskip{}

\textbf{Algorithm for the evaluation of $G(z)$.} 
\begin{enumerate}
\item Find the index sets ${\cal I}_{+}$, ${\cal I}_{-}$, ${\cal I}_{0}$,
and their respective cardinalities $k,h,l$. 
\item If $k=0$, set $G(z)=0$ and exit. 
\item Set $S_{+}=\sum_{i=1}^{k}z_{i}$. 
\item If $h=0$, set $G(z)=\min(p,S_{+})$, and exit. 
\item If $h>0$, order the negative elements of $z$, and evaluate $\underline{G}(z)$
by the recursion~(\ref{eq:rec-underlineG-svm-app}). Set $G(z)=\min(p,S_{+},\underline{G}(z))$
and exit. 
\end{enumerate}
The complexity of evaluating $G(z)$ thus grows in $O(k+h\log h)$,
which is less than $O(d\log d)$, where $d=k+h$ is the number of
non-zero elements in $z$.

\section{Computing $P_{{\rm log}}(\gamma,x)$ via an interior-point method}

We consider the problem~(\ref{eq:P-lo-pb}) which arises with the
logistic loss. We can use a generic interior-point method \cite{BV:04},
and exploit the decomposable structure of the dual function $G_{{\rm log}}$.
The algorithm is based on solving, via a variant of Newton's method,
a sequence of linearly constrained problems of the form \[
\min_{\theta}\:\tau x^{T}\theta+\log(G_{{\rm log}}(\theta)-\gamma)+\sum_{i=1}^{m}\log(-\theta-\theta^{2})~:~z^{T}\theta=0,\]
 where $\tau>0$ is a parameter that is increased as the algorithm
progresses, and the last terms correspond to domain constraints $\theta\in[-1,0]^{m}$.
As an initial point, we can take the point $\theta$ generated by
scaling, as explained in section~\ref{ss:dual-scaling-generic}.
Each iteration of the algorithm involves solving a linear system in
variable $\delta$, of the form $H\delta=h$, with $H$ is a rank-two
modification to the Hessian of the objective function in the problem
above. It is easily verified that the matrix $H$ has a ``diagonal
plus rank-two'' structure, that is, it can be written as $H=D-gg^{T}-vv^{T}$,
where the $m\times m$ matrix $D$ is diagonal and $g,v\in\reals^{m}$
are computed in $O(m)$. The matrix $H$ can be formed, as the associated
linear system solved, in $O(m)$ time. Since the number of iterations
for this problem with two constraints grows as $\log(1/\epsilon)O(1)$,
the total complexity of the algorithm is $\log(1/\epsilon)O(m)$ ($\epsilon$
is the absolute accuracy at which the interior-point method computes
the objective). We note that memory requirements for this method also
grow as $O(m)$.

\section{On thresholding methods for LASSO}

\label{app:thresholding} Sparse classification algorithms may return
a classifier vector $w$ with many small, but not exactly zero, elements.
This implies that we need to choose a thresholding rule to decide
which elements to set to zero. In this section, we discuss an issue
related to the thresholding rule originally proposed for the interior
point method for Logistic algorithm in \cite{boyd_logistic}, and
propose a new thresholding rule.

\paragraph{The KKT thresholding rule.}

Recall that the primal problem for LASSO is \begin{equation}
\phi(\lambda)=\min_{w}\frac{1}{2}\|X^{T}w-y\|_{2}^{2}+\lambda\|w\|_{1}.\end{equation}
 Observing that the KKT conditions imply that, at optimum, $(X(X^{T}w-y))_{k}=\lambda\mbox{sign}(w_{k})$,
with the convention $\mbox{sign}(0)\in[-1,1]$, and following the
ideas of \cite{boyd_logistic}, the following thresholding rule can
be proposed: at optimum, set component $w_{k}$ to $0$ whenever \begin{equation}
|(X(X^{T}w-y))_{k}|\leq0.9999\lambda.\label{thresholding_Boyd_rule}\end{equation}
 We refer to this rule as the ``KKT'' rule.

The IPM-LASSO algorithm takes as input a ``duality gap'' parameter
$\epsilon$, which controls the relative accuracy on the objective.
When comparing the IPM code results with other algorithms such as
GLMNET, we observed chaotic behaviors when applying the KKT rule,
especially when the duality gap parameter $\epsilon$ was not small
enough. More surprisingly, when this parameter is not small enough,
some components $w_{k}$ with absolute values not close to $0$ can
be thresholded. This suggests that the KKT rule should only be used
for problems solved with a small enough duality gap $\epsilon$. However,
setting the duality gap to a small value can dramatically slow down
computations. In our experiments, changing the duality gap from $\epsilon=10^{-4}$
to $10^{-6}$ (resp.\
$10^{-8}$) increased the computational time by $30\%$ to $40\%$
(resp. $50$ to $100\%$).

\paragraph{An alternative method.}

We propose an alternative thresholding rule, which is based on controlling
the perturbation of the objective function that is induced by thresholding.

Assume that we have solved the LASSO problem above, with a given duality
gap parameter $\epsilon$. If we denote by $w^{\ast}$ the classifier
vector delivered by the IPM algorithm, $w^{\ast}$ is $\epsilon$-sub-optimal,
that is, achieves a value \[
\phi^{\ast}=\frac{1}{2}\|Xw^{\ast}-y\|_{2}^{2}+\lambda\|w^{\ast}\|_{1},\]
 with $0\le\phi^{\ast}-\phi(\lambda)\le\epsilon\phi(\lambda)$.

For a given threshold $\tau>0$, consider the thresholded vector $\tilde{w}(\tau)$
defined as \begin{eqnarray*}
\tilde{w}_{k}(\tau) & = & \left\{ \begin{array}{ll}
0 & \mbox{if }|w_{k}^{\ast}|\le\tau,\\
w_{k}^{\ast} & \mbox{otherwise,}\end{array}\right.\;\; k=1,\ldots,n.\end{eqnarray*}
 We have $\tilde{w}(\tau)=w^{\ast}+\delta(\tau)$ where the vector
of perturbation $\delta(\tau)$ is such that \begin{eqnarray*}
\delta_{k}(\tau) & = & \left\{ \begin{array}{ll}
-w_{k}^{\ast} & \mbox{if }|w_{k}^{\ast}|\le\tau,\\
0 & \mbox{otherwise,}\end{array}\right.\;\; k=1,\ldots,n.\end{eqnarray*}
 Note that, by construction, we have $\|w^{\ast}\|_{1}=\|w^{\ast}+\delta\|_{1}+\|\delta\|_{1}$.
Also note that if $w^{\ast}$ is sparse, so is $\delta$.

Let us now denote by $\phi_{\tau}$ the LASSO objective that we obtain
upon replacing the optimum classifier $w^{\ast}$ with its thresholded
version $\tilde{w}(\tau)=w^{\ast}+\delta(\tau)$: \begin{eqnarray*}
\phi_{\tau} & := & \frac{1}{2}\|X(w^{\ast}+\delta(\tau))-y\|_{2}^{2}+\lambda\|w^{\ast}+\delta(\tau)\|_{1}.\end{eqnarray*}

Since $w(\tau)$ is (trivially) feasible for the primal problem, we
have $\phi_{\tau}\ge\phi(\lambda)$. On the other hand, \begin{eqnarray*}
\phi_{\tau} & = & \frac{1}{2}\|Xw^{\ast}-y\|_{2}^{2}+\lambda\|w^{\ast}+\delta(\tau)\|_{1}+\frac{1}{2}\|X\delta(\tau)\|_{2}^{2}+\delta(\tau)^{T}X^{T}(Xw^{\ast}-y)\\
 & \le & \frac{1}{2}\|Xw^{\ast}-y\|_{2}^{2}+\lambda\|w^{\ast}\|_{1}+\frac{1}{2}\|X\delta(\tau)\|_{2}^{2}+\delta(\tau)^{T}X^{T}(Xw^{\ast}-y).\end{eqnarray*}
 For a given $\alpha>1$, the condition \begin{equation}
{\cal C}(\tau):=\frac{1}{2}\|X\delta(\tau)\|_{2}+\delta(\tau)^{T}X^{T}(Xw^{\ast}-y)\leq\kappa\phi^{\ast},\;\;\kappa:=\frac{1+\alpha\epsilon}{1+\epsilon}-1\ge0,\label{eq:new-thresholding-rule}\end{equation}
 allows to write \[
\phi(\lambda)\le\phi_{\tau}\le\ (1+\alpha\epsilon)\phi(\lambda).\]
 The condition~(\ref{eq:new-thresholding-rule}) then implies that
the thresholded classifier is sub-optimal, with relative accuracy
$\alpha\epsilon$.

Our proposed thresholding rule is based on the condition~(\ref{eq:new-thresholding-rule}).
Precisely, we choose the parameter $\alpha>0$, then we set the threshold
level $\tau$ by solving, via line search, the largest threshold $\tau$
allowed by condition~(\ref{eq:new-thresholding-rule}): \[
\tau_{\alpha}=\arg\max_{\tau\geq0}\:\left\{ \tau~:~\|X\delta(\tau)\|_{2}\le\left(\sqrt{\frac{1+\alpha\epsilon}{1+\epsilon}}-1\right)\|Xw^{\ast}-y\|_{2}\right\} .\]
 The larger $\alpha$ is, the more elements the rule allows to set
to zero; at the same time, the more degradation in the objective will
be observed: precisely, the new relative accuracy is bounded by $\alpha\epsilon$.
The rule also depends on the duality gap parameter $\epsilon$. We
refer to the thresholding rule as TR($\alpha$) in the sequel. In
practice, we observe that the value $\alpha=2$ works well, in a sense
made more precise below.

The complexity of the rule is $O(mn)$. More precisely, the optimal
dual variable $\theta^{\ast}=Xw^{\ast}-y$ is returned by IPM-LASSO.
The matrix $X\theta^{\ast}=X(X^{T}w^{\ast}-y)$ is computed once for
all in $O(mn)$. We then sort the optimal vector $w^{\ast}$ so that
$|w_{(1)}^{\ast}|\le\ldots\le|w_{(n)}^{\ast}|$, and set $\tau=\tau_{0}=|w_{(n)}^{\ast}|$,
so that $\delta_{k}(\tau_{0})=-w_{k}^{\ast}$ and $\tilde{w}_{k}(\tau_{0})=0$
for all $k=1,\ldots,n$. The product $X\delta(\tau_{0})$ is computed
in $O(mn)$, while the product $\delta(\tau_{0})^{T}(X^{T}\theta^{\ast})$
is computed in $O(n)$. If the quantity ${\cal C}(\tau_{0})=\frac{1}{2}\|X\delta(\tau_{0})\|_{2}+\delta(\tau_{0})^{T}(X^{T}\theta^{\ast})$
is greater than $\kappa\phi^{\star}$, then we set $\tau=\tau_{1}=|w_{(n-1)}^{\ast}|$.
We have $\delta_{k}(\tau_{1})=\delta_{k}(\tau_{0})$ for any $k\neq(n)$
and $\delta_{(n)}(\tau_{1})=0$. Therefore, ${\cal C}(\tau_{1})$
can be deduced from ${\cal C}(\tau_{0})$ in $O(n)$. We proceed by
successively setting $\tau_{k}=|w_{(n-k)}^{\ast}|$ until we reach
a threshold $\tau_{k}$ such that ${\cal C}(\tau_{k})\le\kappa\phi^{\ast}$.

\paragraph{Simulation study.}

We conducted a simple simulation study to evaluate our proposal and
compare it to the KKT thresholding rule. Both methods were further
compared to the results returned by the \verb"glmnet" \verb"R" package.
The latter algorithm returns hard zeros in the classifier coefficients,
and we have chosen the corresponding sparsity pattern as the ``ground
truth'', which the IPM should recover.

We first experimented with synthetic data. We generated samples of
the pair $(X,y)$ for various values of $(m,n)$. We present the results
for $(m,n)=(5000,2500)$ and $(m,n)=(100,500)$. The number $s$ of
relevant features was set to $\min(m,n/2)$. Features were drawn from
independent ${\mathcal{N}}(0,1)$ distributions and $y$ was computed
as $y=X^{T}w+\xi$, where $\xi\sim{\mathcal{N}}(0,0.2)$ and $w$
is a vector of $\mathbb{R}^{n}$ with first $s$ components equal
to $0.1+1/s$ and remaining $n-s$ components set to $0$. Because
\verb"glmnet" includes an unpenalized intercept while IPM method
does not, both $y$ and $X$ were centered before applying either
methods to make their results comparable.

\begin{figure}
\begin{tabular}{cc}
\includegraphics[width=0.48\textwidth]{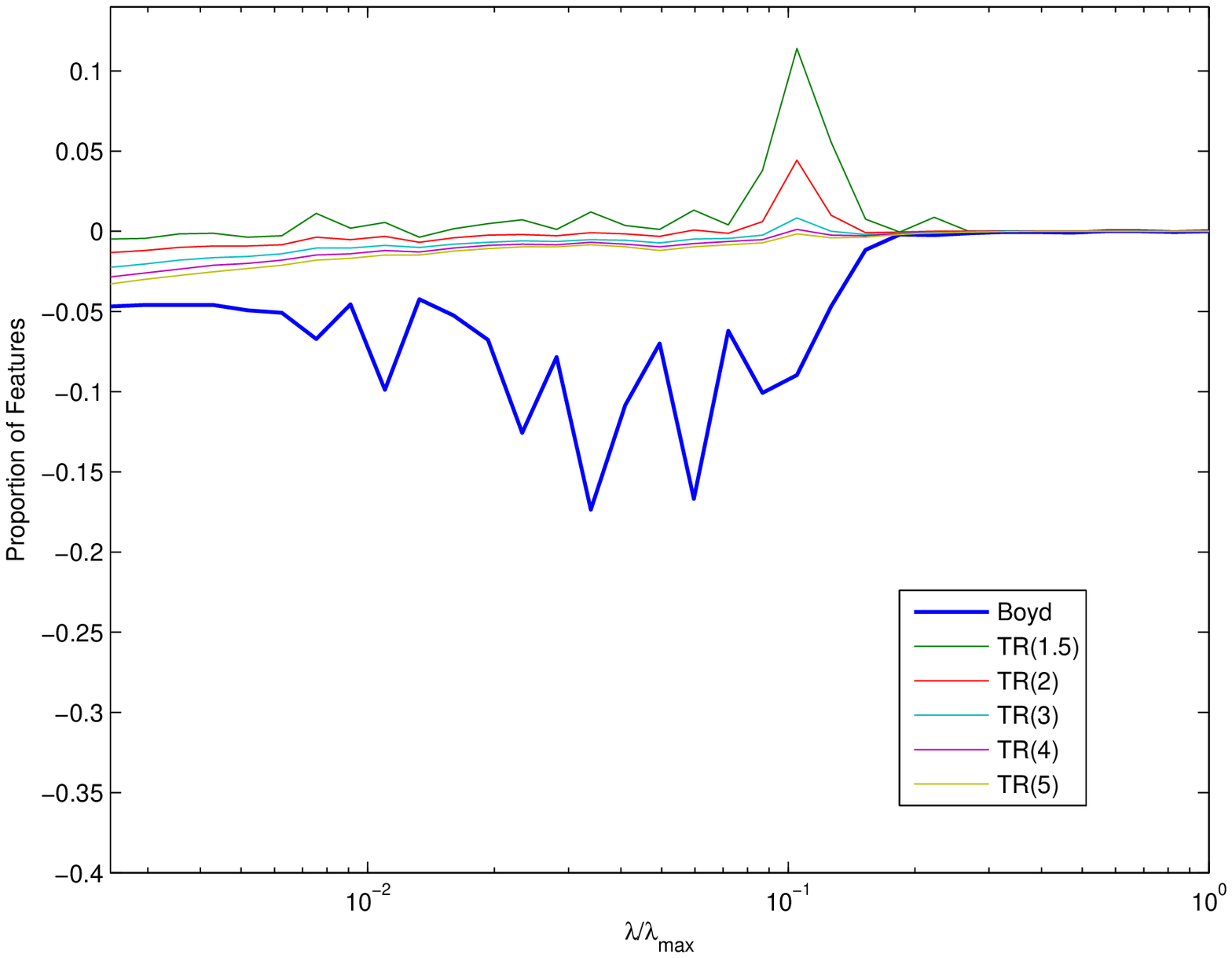}
\includegraphics[width=0.48\textwidth]{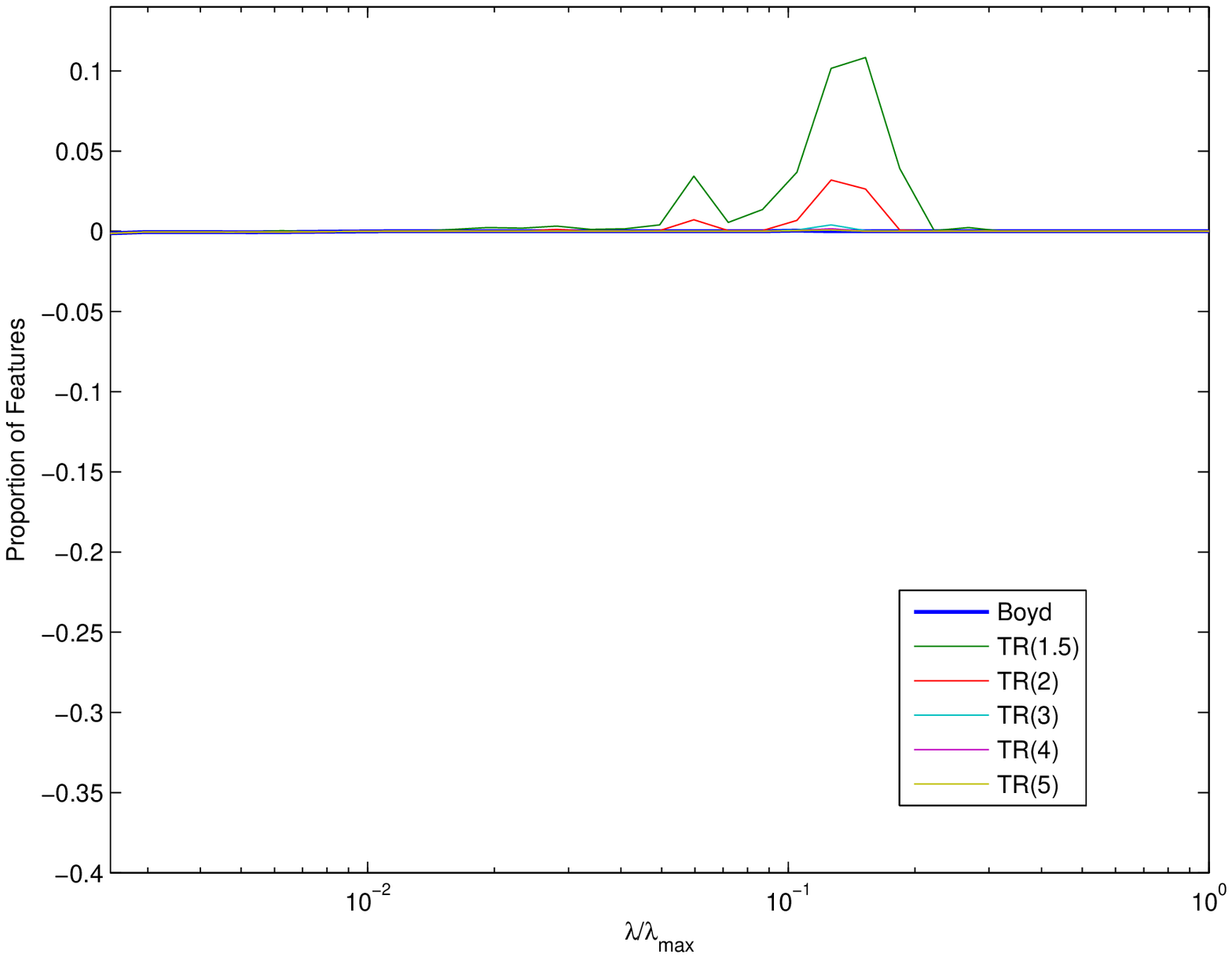} & \tabularnewline
\includegraphics[width=0.48\textwidth]{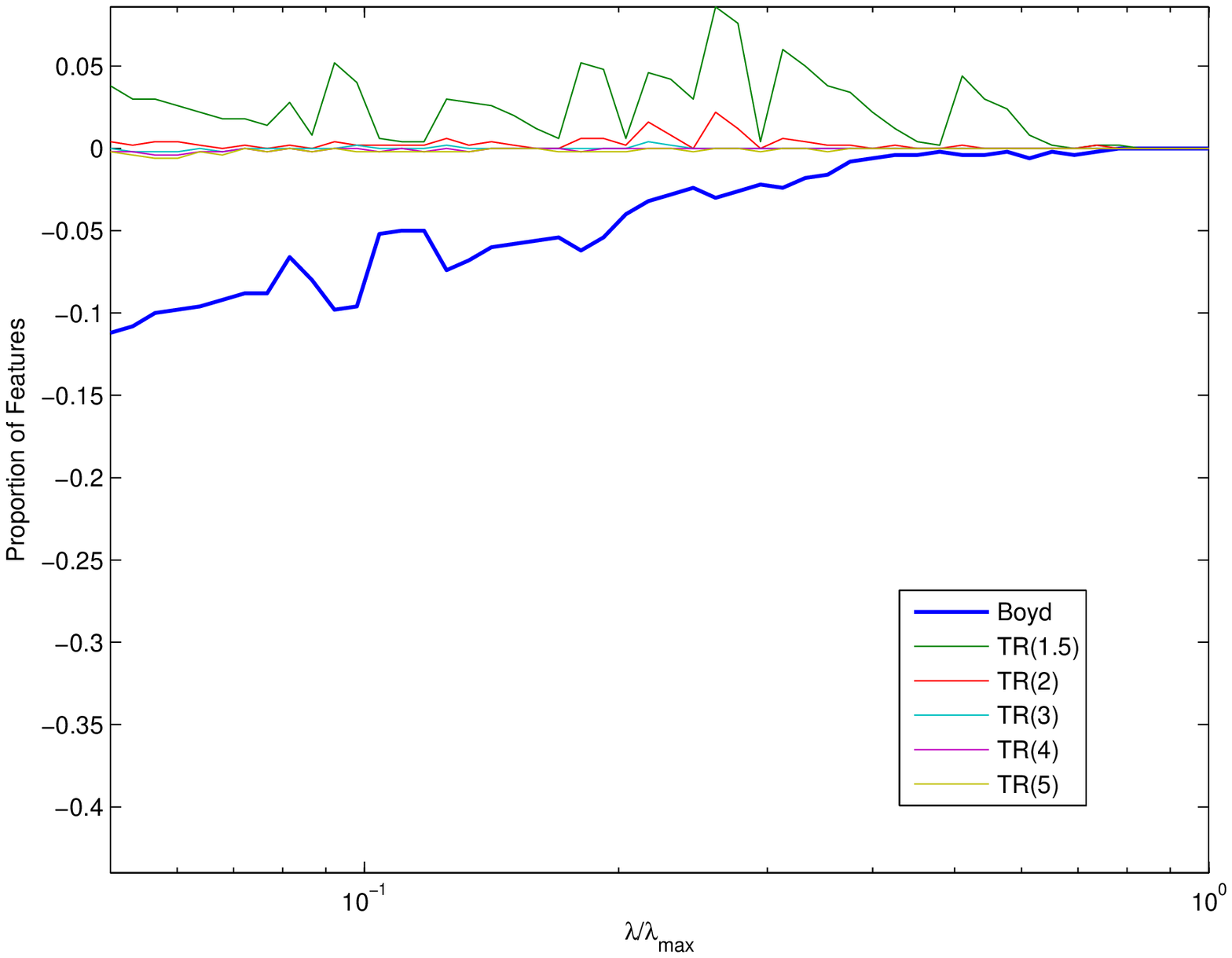}
\includegraphics[width=0.48\textwidth]{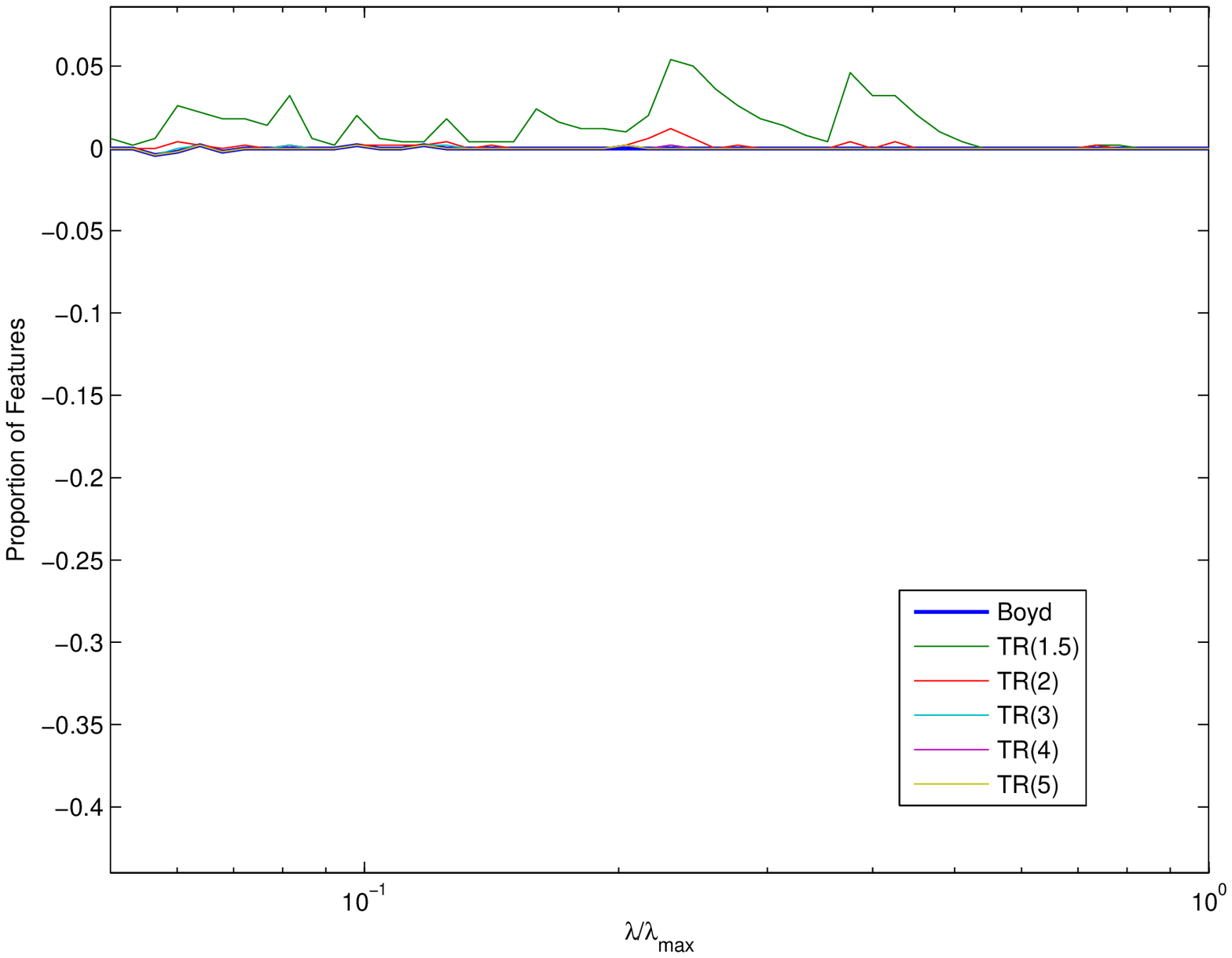}  & \tabularnewline
\end{tabular}\caption{\label{fig:thresholding_Sim_1} Comparison of several thresholding
rules on synthetic data: the case $m=5000$, $n=100$ ({\em top
panel}) and $m=100$, $n=500$ ({\em bottom panel}) with duality
gap in IPM method set to $(i)$ $10^{-4}$ ({\em left panel}) and
$(iii)$ $10^{-8}$ ({\em right panel}). The curves represent the
differences between the number of active features returned after each
thresholding method and the one returned by \texttt{glmnet} (this
difference is further divided by the total number of features $n$).
The graphs present the results attached to six thresholding rules:
the one proposed by \cite{boyd_logistic} and five versions of our
proposal, corresponding to setting $\alpha$ in (\ref{eq:new-thresholding-rule})
to $1.5$, $2$, $3$, $4$ and $5$ respectively. Overall, these
results suggest that by setting $\alpha\in(2,5)$, our rule is less
sensitive to the value of the duality gap parameter in IPM-LASSO than
is the rule proposed by \cite{boyd_logistic}.}

\end{figure}

Results are presented on Figures \ref{fig:thresholding_Sim_1}. First,
the KKT thresholding rule was observed to be very chaotic when the
duality gap was set to $\epsilon=10^{-4}$ (we recall here that the
default value for the duality gap in IPM \texttt{MATLAB} implementation
is $\epsilon=10^{-3}$), while it was way better when duality gap
was set to $\epsilon=10^{-8}$ (somehow justifying our choice of considering
the sparsity pattern returned by \verb"glmnet" as the ground truth).
Therefore, for applications where computational time is not critical,
running IPM method and applying KKT thresholding rule should yield
appropriate results. However, when computational time matters, passing
the duality gap from, say, $10^{-4}$ to $10^{-8}$, is not a viable
option. Next, regarding our proposal, we observed that it was significantly
better than KKT thresholding rule when the duality gap was set to
$10^{-4}$ and equivalent to KKT thresholding rule for a duality gap
of $10^{-8}$. Interestingly, setting $\alpha=1.5$ in (\ref{eq:new-thresholding-rule})
generally enabled to achieved very good results for low values of
$\lambda$, but lead to irregular results for higher values of $\lambda$
(in the case $m=100$, results were unstable for the whole range of
$\lambda$ values we considered). Overall, the choices $\alpha=2$,
$3$ and $4$ lead to acceptable results. A little irregularity remained
with $\alpha=2$ for high values of $\lambda$, but this choice of
$\alpha$ performed the best for lower values of $\lambda$. As for
choices $\alpha=3$ and $\alpha=4$, it is noteworthy that the results
were all the better as the dimension $n$ was low.

\subsection{Real data examples}

\begin{figure}
\begin{tabular}{cc}
\includegraphics[width=0.48\textwidth]{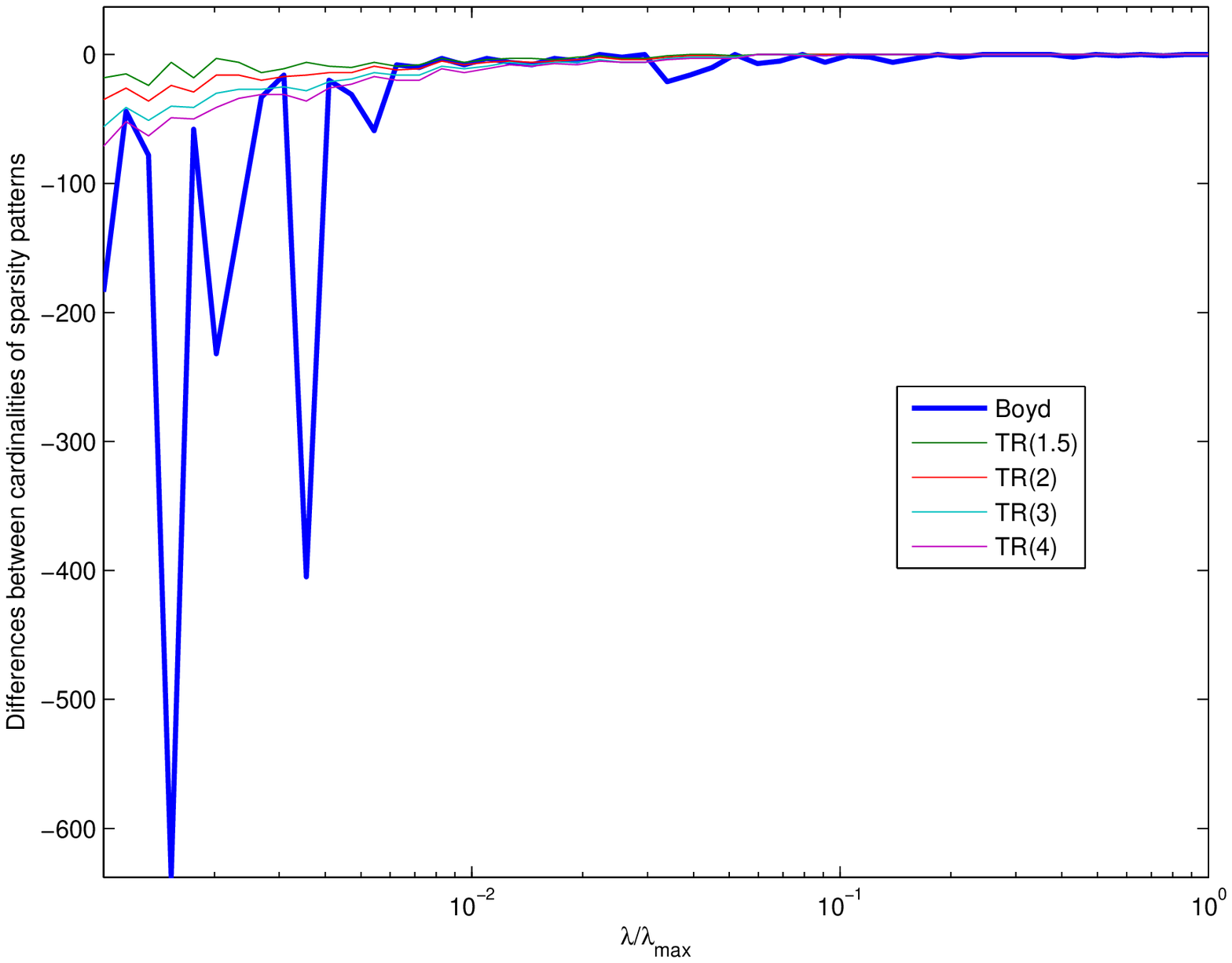}
\includegraphics[width=0.48\textwidth]{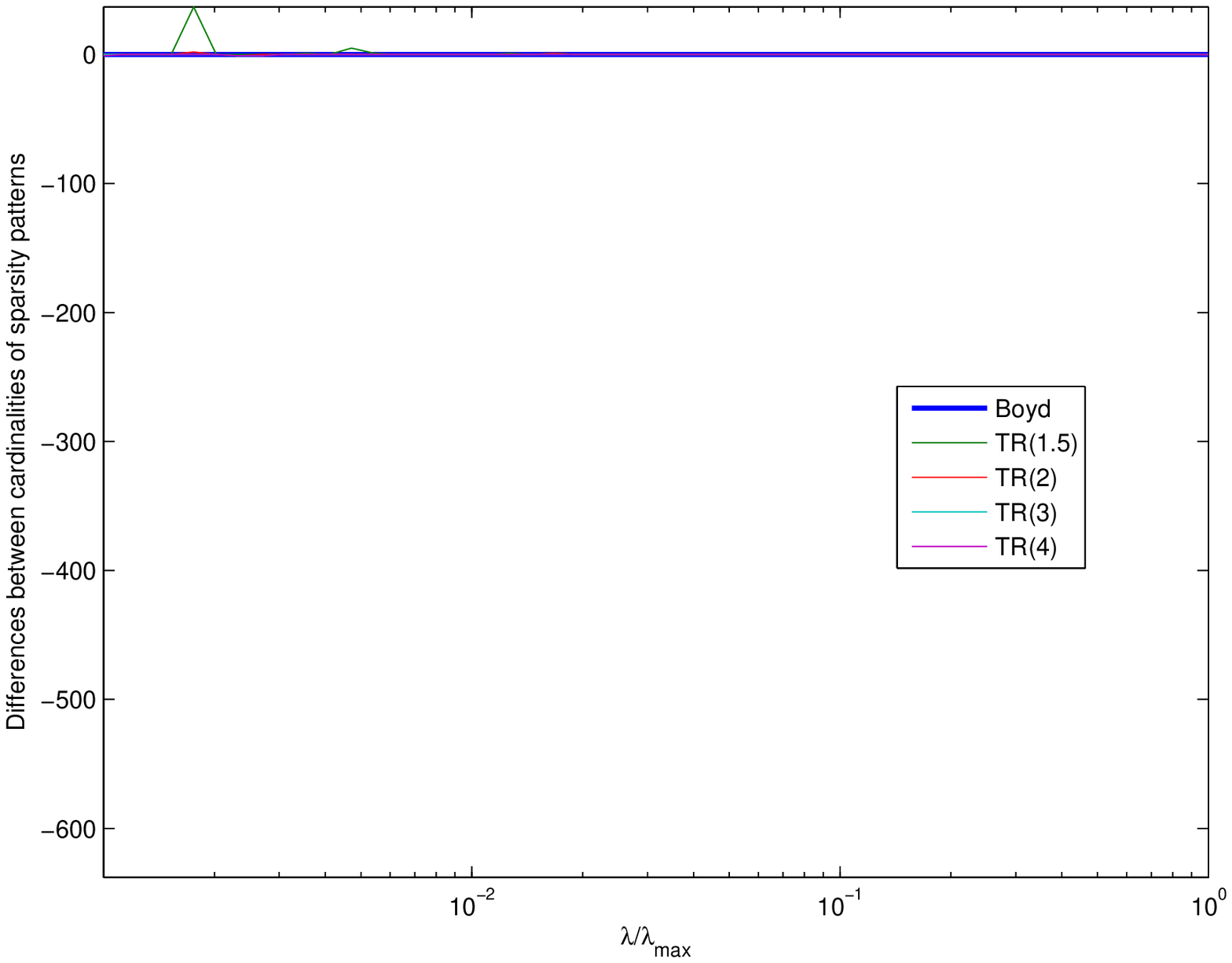}  & \tabularnewline
\end{tabular}\caption{\label{fig:thresholding_NYT_China} Comparison of several thresholding
rules on the NYT headlines data set for the topic \textquotedbl{}China\textquotedbl{}
and year $1985$. Duality gap in IPM-LASSO was successively set to
$10^{-4}$ ({\em left panel}) and $10^{-8}$ ({\em right panel}).
The curves represent the differences between the number of active
features returned after each thresholding method and the one returned
by the KKT rule when duality gap was set to $10^{-10}$. The graphs
present the results attached to five thresholding rules: the KKT rule
and four versions of our rule, corresponding to setting $\alpha$
in (\ref{eq:new-thresholding-rule}) to $1.5$, $2$, $3$ and $4$
respectively. Results obtained following our proposal appear to be
less sensitive to the value of the duality gap used in IPM-LASSO.
For instance, for the value $\lambda=\lambda_{{\rm max}}/1000$, the
KKT rule returns $1758$ active feature when the duality gap is set
to $10^{-4}$ while it returns $2357$ features for a duality gap
of $10^{-8}$.}

\end{figure}

We also applied our proposal and compared it to KKT rule (\ref{thresholding_Boyd_rule})
on real data sets arising in text classification. More precisely,
we used the New York Times headlines data set presented in the Numerical
results Section. For illustration, we present here results we obtained
for the topic \textquotedbl{}China\textquotedbl{} and the year $1985$.
We successively ran IPM-LASSO method with duality gap set to $10^{-4}$
and $10^{-8}$ and compare the number of active features returned
after applying KKT thresholding rule (\ref{thresholding_Boyd_rule})
and TR $(1.5)$, TR $(2)$, TR $(3)$ and TR $(4)$. Results are presented
on Figure \ref{fig:thresholding_NYT_China}. Because we could not
applied \texttt{glmnet} on this data set, the ground truth was considered
as the result of KKT rule, when applied to the model returned by IPM-LASSO
ran with duality gap set to $10^{-10}$. Applying KKT rule on the
model built with a duality gap of $10^{-4}$ lead to very misleading
results again, especially for low values of $\lambda$. In this very
high-dimensional setting ($n=38377$ here), our rule generally resulted
in a slight \textquotedbl{}underestimation\textquotedbl{} of the true
number of active features for the lowest values of $\lambda$ when
the duality gap was set to $10^{-4}$. This suggests that the ``optimal''
$\alpha$ for our rule might depend on both $n$ and $\lambda$ when
the duality gap is not small enough. However, we still observed that
our proposal significantly improved upon KKT rule when the duality
gap was set to $10^{-4}$.

\newpage

\bibliography{ffe_jmlr}

\end{document}